\documentclass{article}
\usepackage[T1]{fontenc}
\usepackage[utf8]{inputenc}

\usepackage[preprint]{ICML2026/icml2026}

\usepackage{graphicx}
\graphicspath{{figures/}}
\usepackage{amsmath, amssymb}
\usepackage{algorithm}
\usepackage{algorithmic}
\usepackage{booktabs}
\usepackage{multirow}
\usepackage{xcolor}
\usepackage{bm}
\usepackage{tikz}
\usepackage{array}
\usepackage{booktabs}
\usepackage{subcaption}
\usepackage[normalem]{ulem}
\usetikzlibrary{arrows.meta, positioning, shapes.geometric, fit}
\usepackage[table]{xcolor}
\definecolor{hitgray}{gray}{0.9} 
\newcommand{\hitcell}[1]{\cellcolor{hitgray}#1}
\usepackage{makecell}
\newcommand{\twolines}[2]{\makecell[r]{#1\\#2}}

\usepackage{natbib}

\definecolor{myblue}{rgb}{0.21,0.49,0.74}
\usepackage[pagebackref,breaklinks,colorlinks,allcolors=myblue]{hyperref}

\icmltitlerunning{LEMON}

\begin{document}

\twocolumn[
\icmltitle{LEMON: Local Explanations via Modality-aware OptimizatioN}

\begin{icmlauthorlist}
  \icmlauthor{Yu Qin}{bristol}
  \icmlauthor{Phillip Sloan}{bristol}
  \icmlauthor{Raul Santos-Rodriguez}{bristol}
  \icmlauthor{Majid Mirmehdi}{bristol}
  \icmlauthor{Telmo de Menezes e Silva Filho}{bristol}
\end{icmlauthorlist}

\icmlaffiliation{bristol}{University of Bristol, Bristol, United Kingdom}

\icmlcorrespondingauthor{Yu Qin}{gw21127@bristol.ac.uk}
\icmlcorrespondingauthor{Telmo de Menezes e Silva Filho}{telmo.silvafilho@bristol.ac.uk}

\vskip 0.3in
]

\printAffiliationsAndNotice{}

\begin{abstract}
\label{ref:abstract_icml}
Multimodal models are ubiquitous, yet existing explainability methods are often single-modal, architecture-dependent, or too computationally expensive to run at scale. We introduce LEMON (Local Explanations via Modality-aware OptimizatioN), a model-agnostic framework for local explanations of multimodal predictions. LEMON fits a single modality-aware surrogate with group-structured sparsity to produce unified explanations that disentangle modality-level contributions and feature-level attributions. The approach treats the predictor as a black box and is computationally efficient, requiring relatively few forward passes while remaining faithful under repeated perturbations. We evaluate LEMON on vision–language question answering and a clinical prediction task with image, text, and tabular inputs, comparing against representative multimodal baselines. Across backbones, LEMON achieves competitive deletion-based faithfulness while reducing black-box evaluations by 35--67$\times$ and runtime by 2--8$\times$ compared to strong multimodal baselines. Code will be made available.
\end{abstract}

\section{Introduction}
\label{sec:introduction}
Deep learning models have achieved impressive performance across a wide range of tasks involving heterogeneous data sources, including images, text, signals, and structured tables. However, their inherent complexity presents significant challenges for interpretability, especially in high-stakes domains, such as autonomous systems \citep{tian2024drivevlm}, finance \citep{cerneviciene2024xai_finance, wilson2025xai_finance}, and healthcare \citep{thirunavukarasu2023medicine}, where transparency is critical for informed decision making. A lack of explanation can undermine trust among domain experts and end-users, thereby limiting the adoption of otherwise effective AI systems \citep{klein2024finance, rosenbacke2024xai_trust, judijanto2025trust}. The necessity for explainability is exemplified by complex tasks such as differential medical diagnosis, where multiple conditions may be compatible with the same high-level findings and subtle, modality-specific cues are required to distinguish between them \citep{ma2024ftd_differential}.

\begin{figure}[!h]
  \centering
  \includegraphics[width=\linewidth]{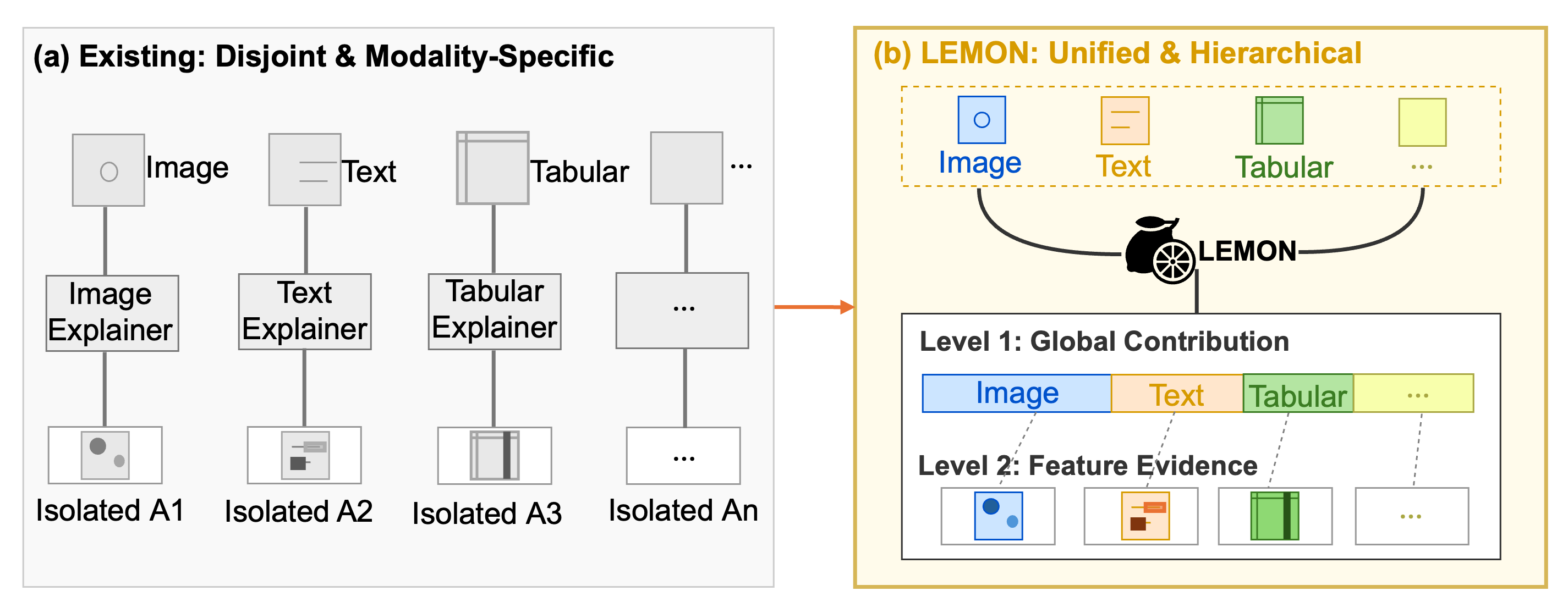}
  \caption{\textbf{Conceptual overview of LEMON.} LEMON queries a black-box model with group-structured perturbations and fits a Sparse Group Lasso surrogate to output a two-level explanation: modality contributions and within-modality evidence.}
  \label{fig:lemon-concept}
\end{figure}

While modern multimodal architectures, often transformer-based, such as CLIP \citep{radford2021clip} and LLaVA \citep{liu2023llava}, can effectively integrate multiple modalities, interpretability methods have not kept pace \citep{dang2024gap}. Most existing approaches remain modality-specific: they either handle only unimodal inputs or produce isolated per-modality explanations that can be misleading in multimodal settings~\citep{agarwal2025rethink}. This fragmented design obscures how contributions are distributed across and within modalities, making it difficult to discern whether a prediction arises from single-modality dominance or genuinely multi-source reasoning. Furthermore, many model-agnostic post-hoc methods rely on stochastic perturbation or coalition sampling, which can require repeated black-box evaluations and may exhibit non-trivial run-to-run variability in their attributions \citep{jethani2022fastshap, kelodjou2024shaping}. The lack of structured explanations that simultaneously capture modality- and feature-level effects, while remaining efficient and stable, remains a significant limitation in current multimodal interpretability practice.

This limitation has practical consequences. Undesirable behaviors such as unimodal collapse \citep{parcalabescu2023mmshap}, where a model disproportionately relies on one modality while ignoring others, may not reduce overall predictive accuracy and therefore remain undetected by standard evaluation metrics. Existing interpretability tools offer only limited means to quantify such asymmetric reliance or to relate it to concrete input features. These gaps motivate three questions that guide our study: {(i)} can a local explainer reveal systematic differences in explanation behavior between an untrained model and a trained model on the same task; {(ii)} how does a structured, model-agnostic explainer compare with representative multimodal baselines under matched evaluation budgets; and {(iii)} do the resulting explanations align with ground-truth evidence in a multimodal clinical application?

In this work, we propose {LEMON} (\textbf{L}ocal \textbf{E}xplanations via \textbf{M}odality-aware \textbf{O}ptimizatio\textbf{N}), a multimodal extension of the popular LIME explainer \citep{ribeiro2016lime} tailored for model-agnostic local explanations. As shown in Figure \ref{fig:lemon-concept}, LEMON fits a single surrogate model with sparse group-lasso regularization \citep{simon2013sparsegrouplasso} that jointly captures modality-level contributions and within-modality feature importances in a single optimization problem. The framework is both model-agnostic and data-agnostic, interacting with the black-box predictor only through input–output queries and without requiring access to gradients, attention weights, or other internal signals. We instantiate LEMON on models with image–text and image–text–tabular inputs, using standard partitioners to define interpretable components. The use of a group-sparse surrogate keeps query complexity controlled, yielding compact explanations that are comparatively inexpensive to compute and stable under a matched query budget.

Our contributions are threefold. {(i)} We reformulate local surrogate modelling for multimodal predictors using a \emph{single} sparse, group-structured surrogate, producing explicitly hierarchical explanations that quantify both modality contributions and feature effects within each modality. {(ii)} We introduce a simple, partitioner-agnostic interface that decouples the explainer from the definition of interpretable components, enabling different image and text partitioning strategies to be plugged into the same optimization procedure. {(iii)} We provide diagnostic use cases and a comprehensive empirical evaluation: we use LEMON to audit modality reliance and identify salient evidence for individual predictions, and we benchmark on synthetic data with known ground-truth attributions and on real-world multimodal tasks, comparing against representative model-agnostic baselines.

\section{Related Work}
\label{sec:relatedwork}
\begin{table*}[t]
\centering
\caption{\textbf{Comparison of representative interpretability methods.} Here, \emph{unified hierarchical} means a single explanation model yielding modality- and feature-level attributions; \emph{modality-flexible} means not tied to a fixed modality set; and \emph{practical} means feasible under a matched query budget without subset enumeration or training an explainer.}
\label{tab:comparison}
\resizebox{\textwidth}{!}{%
\begin{tabular}{l|ccccc}
\toprule
\textbf{Method} & \textbf{Model-agnostic} & \textbf{Multimodal} & \textbf{Unified hierarchical} & \textbf{Modality-flexible} & \textbf{Practical (matched budget)} \\
\midrule\midrule
\textit{Model-intrinsic} &  &  &  &  &  \\
Gradient~\citep{selvaraju2017gradcam} & \texttimes & \checkmark & \texttimes & \texttimes & \checkmark \\

\midrule
\multicolumn{6}{l}{\textit{Post-hoc: Shapley-value (SHAP-style)}}\\
SHAP~\citep{lundberg2017shap}      & \checkmark & \texttimes & \texttimes & \checkmark & \texttimes \\
MM\textnormal{-}SHAP~\citep{parcalabescu2023mmshap}   & \checkmark & \checkmark & \texttimes & \texttimes & \texttimes \\
C-SHAP~\citep{Ranjbaran2025cshap}  & \checkmark & \texttimes & \texttimes & \checkmark & \texttimes \\
MultiSHAP~\citep{wang2025multishap} & \checkmark & \checkmark & \texttimes & \texttimes & \texttimes \\

\midrule
\multicolumn{6}{l}{\textit{Post-hoc: learned explainer}}\\
CXPlain~\citep{schwab2019cxplain} & \checkmark & \texttimes & \texttimes & \checkmark & \texttimes \\

\midrule
\multicolumn{6}{l}{\textit{Post-hoc: local surrogate (LIME-style)}}\\
LIME~\citep{ribeiro2016lime}      & \checkmark & \texttimes & \texttimes & \checkmark & \checkmark \\
DIME~\citep{lyu2022dime}      & \checkmark & \checkmark & \texttimes & \texttimes & \texttimes \\
MusicLIME~\citep{sotirou2025musiclime} & \checkmark & \checkmark & \texttimes & \texttimes & \checkmark \\
\textbf{LEMON} & \checkmark & \checkmark & \checkmark & \checkmark & \checkmark \\
\bottomrule
\end{tabular}%
}
\end{table*}

The field of eXplainable AI (XAI) has produced a range of methods to interpret complex models \citep{zhao2023survey}. In the context of multimodal learning, these  can be broadly categorized as \textit{model-intrinsic} and \textit{post-hoc model-agnostic} approaches. LEMON falls into the latter category.
Table~\ref{tab:comparison} provides a compact feature-level comparison.

{\bf Model-Intrinsic Interpretability --} Such methods generate explanations by exploiting internal components of the predictive model, most notably gradient-based signals and attention weights.

\textit{Gradient-based methods}, such as Grad-CAM~\citep{selvaraju2017gradcam}, compute gradients of the model output with respect to internal feature maps to generate saliency visualizations. When applied to multimodal architectures with dedicated visual encoders, they can highlight image regions relevant to the prediction. However, these approaches are tied to specific architectures, and they typically yield separate saliency maps per modality, offering limited insight into how modalities jointly contribute to a prediction.  \textit{Attention-based methods} are widely used in Transformer-based architectures, including many vision–language models such as ViLBERT~\citep{lu2019vilbert} and LXMERT~\citep{tan2019lxmert}, and have been explicitly adapted for multimodal explanation~\citep{chefer2021generic}, where cross-modal attention can be visualized to inspect image–text interactions. The interpretive validity of attention weights has nevertheless been questioned, as high attention scores do not necessarily indicate causal importance~\citep{lopardo2024attention, lyu2024nlp}. Moreover, both attention- and gradient-based techniques require access to internal representations and are thus not suitable as general-purpose tools for auditing black-box multimodal systems.


{\bf Post-Hoc Model-Agnostic Explanations --} 
Post-hoc methods treat the predictor as a black box and explain a prediction through input--output queries.

LIME explains an individual prediction by fitting a simple surrogate---typically a sparse linear model---to perturbed samples around the instance. While effective and scalable, its feature space is typically \emph{flat} (e.g., superpixels, tokens), so in multimodal settings modality-level importance is not explicitly modelled and is often recovered only via post-hoc aggregation. Several multimodal adaptations build on this paradigm: DIME~\citep{lyu2022dime} wraps LIME with a decomposition into unimodal contributions and multimodal interactions, yielding more disentangled views but inheriting feature-level limitations and introducing additional query overhead. MusicLIME similarly instantiates LIME-style local surrogates with domain-specific interpretable components and perturbations in a fixed multimodal music setting~\citep{sotirou2025musiclime}, demonstrating practicality but without a modality-flexible, unified hierarchical surrogate. 

SHAP~\citep{lundberg2017shap} attributes importance via Shapley values under a masking coalition game, but can become expensive in high-dimensional interpretable spaces due to the need to approximate contributions over many coalitions. Efficiency-oriented variants such as C-SHAP (Clustered SHAP) reduce the number of evaluations by clustering background samples for tabular settings~\citep{Ranjbaran2025cshap}, yet it remains unimodal and does not resolve the multimodal partition cost under a matched query budget. MM-SHAP~\citep{parcalabescu2023mmshap} adapts Shapley-style attribution to vision--language models and can be useful for auditing modality reliance, but it typically requires substantially more queries than fitting a single local surrogate and obtains modality-level signals mainly through aggregation rather than an explicitly structured surrogate. MultiSHAP further targets cross-modal interactions~\citep{wang2025multishap}, which is complementary to our focus on unified first-order structured explanations and often increases query cost. Recent generative approaches such as DiffExplainer~\citep{pennisi2024diffexplainer} instead aim at global cross-modal explanations via diffusion models, complementing local surrogate-based auditing but requiring training an explainer.

Causality-inspired methods, such as CXPlain~\citep{schwab2019cxplain}, learn auxiliary models to estimate the performance drop when features are removed, interpreting the estimated effect as a contribution score. 
Such approaches are typically studied in unimodal settings and require training an additional explainer, introducing non-trivial overheads. 

Overall, existing post-hoc pipelines make different compromises for multimodal auditing. LIME-style surrogates are practical but typically yield flat feature-level attributions, so modality-level conclusions are obtained only after aggregation; Shapley-based multimodal variants (e.g., MM-SHAP) can quantify modality usage but often require substantially more queries due to Shapley-style sampling; and wrapper or learned-explainer approaches (e.g., DIME, CXPlain) introduce additional decomposition or training overheads. These trade-offs motivate a model-agnostic approach that produces unified hierarchical explanations from a single sparsity-controlled surrogate under a matched query budget, capturing both modality contributions and within-modality evidence, which is the focus of LEMON.
\section{Methodology}
\label{sec:methodology}
\begin{figure*}[t]
    \centering
    \includegraphics[width=\textwidth]{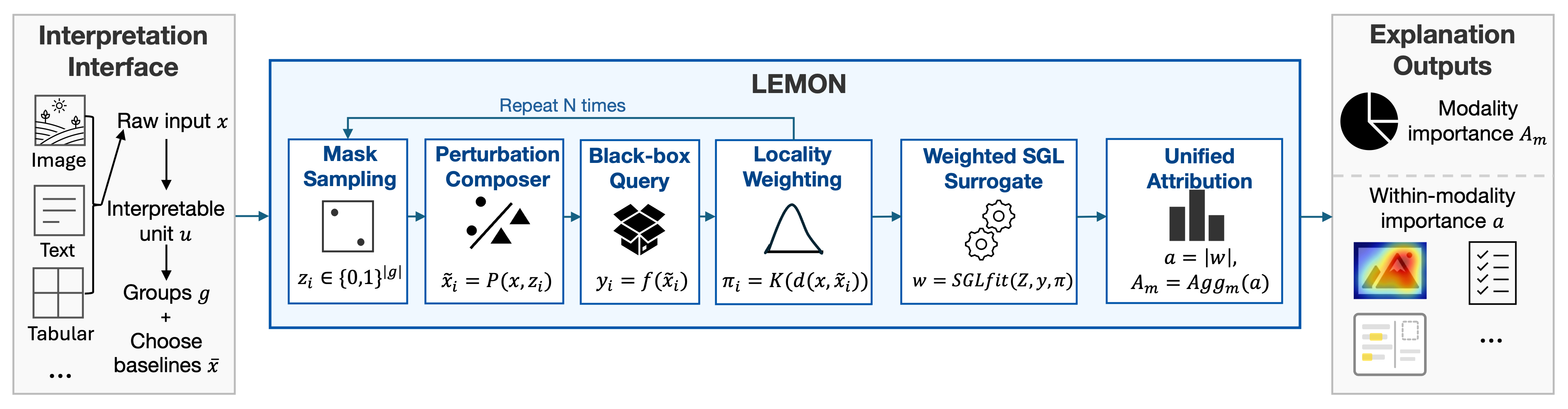}
    \caption{\textbf{Overview of the LEMON framework.} 
    The pipeline generates model-agnostic explanations for multimodal inputs. 
    The input sample is partitioned into interpretable units to preserve modality structures. 
    By employing a \textit{Sparse Group Lasso (SGL)} surrogate, LEMON enforces group-wise sparsity to identify key modalities and regions; see Section~\ref{ssec:sgl} for details. 
    The final outputs include visual heatmaps and text highlights that are evaluated for faithfulness, compactness and cost.}
    
    \label{fig:framework_overview}
\end{figure*}

LEMON is a general-purpose framework for generating hierarchical post-hoc explanations for predictions made by multimodal black-box models. It is designed to be fully model-agnostic and to expose, for each prediction, both (i) modality-level contributions and (ii) within-modality feature attributions, under an explicit query budget. Figure~\ref{fig:framework_overview} summarizes the end-to-end pipeline: an interpretation interface defines interpretable units and modality groups; we then build a local weighted dataset by repeatedly sampling perturbations and querying the black-box; finally, a single weighted Sparse Group Lasso surrogate yields unified attributions.

Formally, let $f$ be a trained multimodal predictor, and let $x=(x^{(1)},\dots,x^{(M)})$ be an instance with $M$ modalities, with $f(x)\in\mathbb{R}^C$. We explain a target scalar output $y(x)$, using a binary interpretable representation and a structured sparse surrogate.

\subsection{Interpretation Interface: Interpretable Units, Groups, and Baselines}
\label{ssec:partitioning}
LEMON starts by mapping each raw modality input into a set of interpretable primitive components.
For each modality $m\in\{1,\dots,M\}$, a user-defined \emph{partitioner} $p_m$ decomposes $x^{(m)}$ into interpretable components $\{x_k^{(m)}\}_{k=1}^{K_m}$, where the $x^{(m)}_k$ are units that can be meaningfully “turned on or off” in perturbations (Figure~\ref{fig:framework_overview}, left).

In principle, $p_m$ can be any modality-specific procedure that yields interpretable units. For images, this could be a region segmentation method such as superpixel or patch-based regions; for text, it could be tokenization into subword tokens, words or phrases; and for temporal signals, it could segement the input into fixed-length windows or event-aligned intervals, and so on. The key requirement is that the resulting units admit a consistent binary masking interface so that LEMON can construct perturbations and fit a surrogate in the shared interpretable space.

After obtaining units from all modalities, we aggregate them into a single interpretable representation.
Let $K=\sum_{m=1}^{M} K_m$ and index all units globally by $\{1,\dots,K\}$. We define modality groups $g_m\subseteq\{1,\dots,K\}$, $m=1,\dots,M$, where $g_m$ indexes components from modality $m$. 
Each perturbation is then represented by a binary interpretable vector $z\in\{0,1\}^K$ over all components, with modality groups $\{g_m\}_{m=1}^M$.
This group structure is passed to the sparse group-lasso regularizer in Section~\ref{ssec:sgl} and is the key mechanism through which LEMON enforces a modality–feature hierarchy.

Finally, the interface specifies a modality-specific baseline (reference) $\bar x^{(m)}$ that instantiates masking in raw input space, enabling the perturbation operator $P(\cdot,\cdot)$ to replace masked units with baseline content while preserving the original input’s format and modality structure.

\subsection{Local Neighborhood Construction}
\label{ssec:neighborhood}
To probe the local behavior of $f$ around $x$, LEMON repeats the following procedure $N$ times (Figure~\ref{fig:framework_overview}, middle): sample a binary mask, compose a perturbed input, query the black-box, and assign a locality weight.

\noindent\textbf{Mask sampling and perturbation composition.}
We sample $z_i\in\{0,1\}^K$ and form a perturbed input $\tilde x_i \;=\; P(x,z_i)$, where $P(\cdot)$ applies modality-specific masking based on $\bar x$ while sharing the same binary representation $z_i$ across modalities.

\noindent\textbf{Black-box query and target transformation.}
We query the predictor once per perturbation to obtain $y_i \;=\; \phi\!\left(f(\tilde x_i)\right)$, where $\phi(\cdot)$ selects and (optionally) transforms the target component to improve local linearity (e.g., logits for classification). This step makes query complexity explicit: LEMON requires $N$ black-box evaluations for perturbed inputs. (Figure~\ref{fig:framework_overview}, black-box query).

\noindent\textbf{Modality-aware locality weights.}
We weight each sample by a kernel $\pi_i$ that downweights perturbations far from the original input. Let $z^{(0)}=\mathbf{1}$ denote the all-ones mask. For each modality group $g_m$, define the per-modality Hamming distance
\begin{equation}
\label{eq:dm_def}
d_m(z) \;=\; \frac{1}{|g_m|}\sum_{j\in g_m}\mathbf{1}\!\left[z_j \neq z^{(0)}_j\right],
\end{equation}
and an IQR-scaled version
\begin{equation}
\label{eq:dm_norm}
\tilde d_m(z) \;=\; \frac{d_m(z)}{\operatorname{IQR}_{i=1,\dots,N}\, d_m(z_i) + \epsilon_d},
\end{equation}
where $\epsilon_d>0$ is a small constant. We aggregate distances with modality weights $\alpha_m$:
\begin{equation}
\label{eq:lemon_kernel}
\begin{aligned}
D(z) &= \sum_{m=1}^{M}\alpha_m\,\tilde d_m(z), \\
\sigma &= \max\!\left(\operatorname{median}_{i=1,\dots N} D(z_i),\,\varepsilon\right), \\
\pi(z) &= \exp\!\left(-\frac{D(z)^2}{\sigma^2}\right) ,
\end{aligned}
\end{equation}
and set $\pi_i=\pi(z_i)$. We use uniform $\alpha_m=1/M$ by default; an optional per-instance selection procedure is deferred to Appendix~\ref{app:alpha}.

Overall, this yields a local weighted dataset
\begin{equation}
\label{eq:local_dataset}
\mathcal{D}=\{(z_i, y_i, \pi_i)\}_{i=1}^N,
\end{equation}
matching the “repeat $N$ times” construction in Figure~\ref{fig:framework_overview}.

\subsection{Weighted Sparse Group Lasso Surrogate}
\label{ssec:sgl}
Given the neighbourhood $\mathcal{D}$ and locality weights $\{\pi_i\}$, LEMON fits a single \emph{weighted} sparse linear surrogate $g(z)=\beta_0+z^\top\beta$ in the interpretable binary space.
Crucially, $\beta$ is structured by modality groups $\{g_m\}_{m=1}^M$ so that the same optimization simultaneously learns modality-level and feature-level effects (Figure~\ref{fig:framework_overview}, weighted SGL surrogate). We solve ~\citep{simon2013sparsegrouplasso}:
\begin{equation}
\label{eq:lemon_obj}
\begin{aligned}
\min_{\beta_0,\beta}\quad
&\frac{1}{2N}\sum_{i=1}^{N} \pi_i\big(y_i-(\beta_0+z_i^\top\beta)\big)^2 \\
&+\lambda(1-\rho)\sum_{m=1}^{M}\tau_m\|\beta_{g_m}\|_2
+\lambda\rho\|\beta\|_1 ,
\end{aligned}
\end{equation}
where $\lambda>0$ controls the overall regularisation, $\rho\in[0,1]$ trades off group vs.\ within-group sparsity, and $\tau_m\geq0$ is the group penalty weight. We apply a simple size-and-variance correction for $\tau_m$ to avoid favoring large or low-variance groups (Appendix~\ref{app:weights}).

The group term encourages entire modalities to be selected or discarded together and the element-wise term yields sparsity within selected modalities. Thus, 
this single optimization problem  produces a modality-feature hierarchy without requiring multiple surrogates or posthoc aggregation.

\subsection{Unified Hierarchical Attribution and Outputs}
\label{ssec:attrib}
We standardize columns of the design matrix under the locality weights $\{\pi_i\}$ (weighted mean zero, unit weighted variance), so coefficient magnitudes are comparable. We define within-modality feature importance as $a_j=|\beta_j|,\quad j=1,\dots,K,$
and modality importance via group aggregation: $A_m=\|\beta_{g_m}\|_2,\quad m=1,\dots,M.$ We rank modalities by $A_m$ and rank units within each modality by $a_j$, which directly drives the explanation outputs (Figure~\ref{fig:framework_overview}, right), such as image heatmaps and text highlights.

\section{Experiments}
\label{sec:experiments}
We design our experiments around three research questions (RQs).\textbf{ RQ1}: does fine-tuning shift the evidence used by CLIP to perform VQA? \textbf{RQ2}: does LEMON outperform existing multimodal explainers according to several metrics? \textbf{RQ3}: do LEMON attributions reflect radiologist-provided diagnostic evidence for different conditions? 
Implementation details, hyperparameter settings, and hardware specifications are provided in Appendix~\ref{app:implementation}.

\subsection{Datasets and tasks}
\label{ssec:datasets}

We evaluate on two public benchmarks: (i) \textbf{VQA v2}~\citep{goyal2017vqa2} validation split for vision--language question answering in RQ1--RQ2, and (ii) \textbf{REFLACX}~\citep{Bigolin2022reflacx} validation subset for clinical evidence alignment with radiologist ellipse annotations in RQ3.

\subsection{Models (black-boxes)}
\label{ssec:models}

\noindent\textbf{CLIP.}
For VQA, we treat CLIP~\citep{radford2021clip} as a black-box image--text scoring model: given an image and question concatenated with a candidate answer as text prompt, it outputs a real-valued compatibility score.

\noindent\textbf{LXMERT.}
We also evaluate on LXMERT~\citep{tan2019lxmert}, a transformer-based vision--language model that takes the question and image as input and produces answer scores. We explain the score of the target answer as a scalar output.

\noindent\textbf{CaMCheX.}
For the clinical task, we use CaMCheX~\citep{sloan2025camchex}, a multimodal model that typically processes full imaging studies, but here is restricted to a single chest x-ray together with clinical text and vitals to output abnormality probabilities. We explain the logit for a target finding, and evaluate against REFLACX ellipse annotations.

\subsection{Evaluation metrics}
\label{ssec:metrics}

\noindent\textbf{Common metrics (RQ1--RQ3).}
We evaluate explanations along four general metrics.
\textbf{Faithfulness} is measured by deletion/insertion AOPC: we progressively remove (or add) features in descending importance order and compute the area over the resulting output curve.
\textbf{Compactness} is measured by $L_0$, the number of selected features.
\textbf{Stability} is measured by Spearman rank correlation of feature importance scores across repeated runs (higher is better).
\textbf{Cost} is reported as wall-clock time per explanation; when available, we additionally report the number of black-box forward calls required to \emph{generate} an explanation, excluding the forward calls used for metric computation.

\noindent\textbf{Clinical alignment metrics (RQ3 only).}
On REFLACX, we assess whether LEMON’s \emph{positive image evidence} aligns with radiologist ellipse annotations on the frontal CXR.
We report complementary localization summaries: (i) \textbf{CH-$z$}, a contrast-heat z-score against a roll-based null; (ii) \textbf{PixelAP$^+$}, pixel-level average precision using the heatmap as a score and the ellipse mask as labels; (iii) \textbf{IoU-AUC$^+$}, the area under a percentile-threshold IoU curve; (iv) \textbf{GT-Cov@K} and \textbf{MassPrec@K}, top-$K$ positive-superpixel coverage/precision decompositions; and (v) \textbf{SPG hit@$\tau$}, a binary hit criterion based on top-$K$ positive superpixels.
All alignment metrics are computed on the image modality and use the same ellipse union for each (study, finding) pair.
In parallel, we report \textbf{modality shares} (image/text/vitals) as the fraction of surrogate coefficient mass per modality to characterise multimodal reliance.

\subsection{RQ1: Inspecting Evidence Shifts in Multimodal VQA Explanations}
LEMON enables qualitative inspection of how model adaptation may affect the multimodal evidence underlying VQA answer scores. In practice, adaptation can plausibly shift explanations from broadly salient image regions or generic prompt tokens toward more question-relevant keywords and their corresponding visual evidence. Such qualitative shifts are consistent with stronger task-specific grounding, suggesting that LEMON can help reveal changes in how a predictor balances and localizes multimodal evidence.

\subsection{RQ2: Comparison to Multimodal Explanation Baselines}
\begin{figure}[t]
  \centering
  \noindent
  \begin{minipage}[c]{0.04\linewidth}
  \centering
    {\small\bfseries\rotatebox{90}{CLIP}}
  \end{minipage}\hfill
  \begin{minipage}[c]{0.95\linewidth}
    \centering
    \begin{subfigure}[b]{0.32\linewidth}
      \centering
      \includegraphics[width=\linewidth]{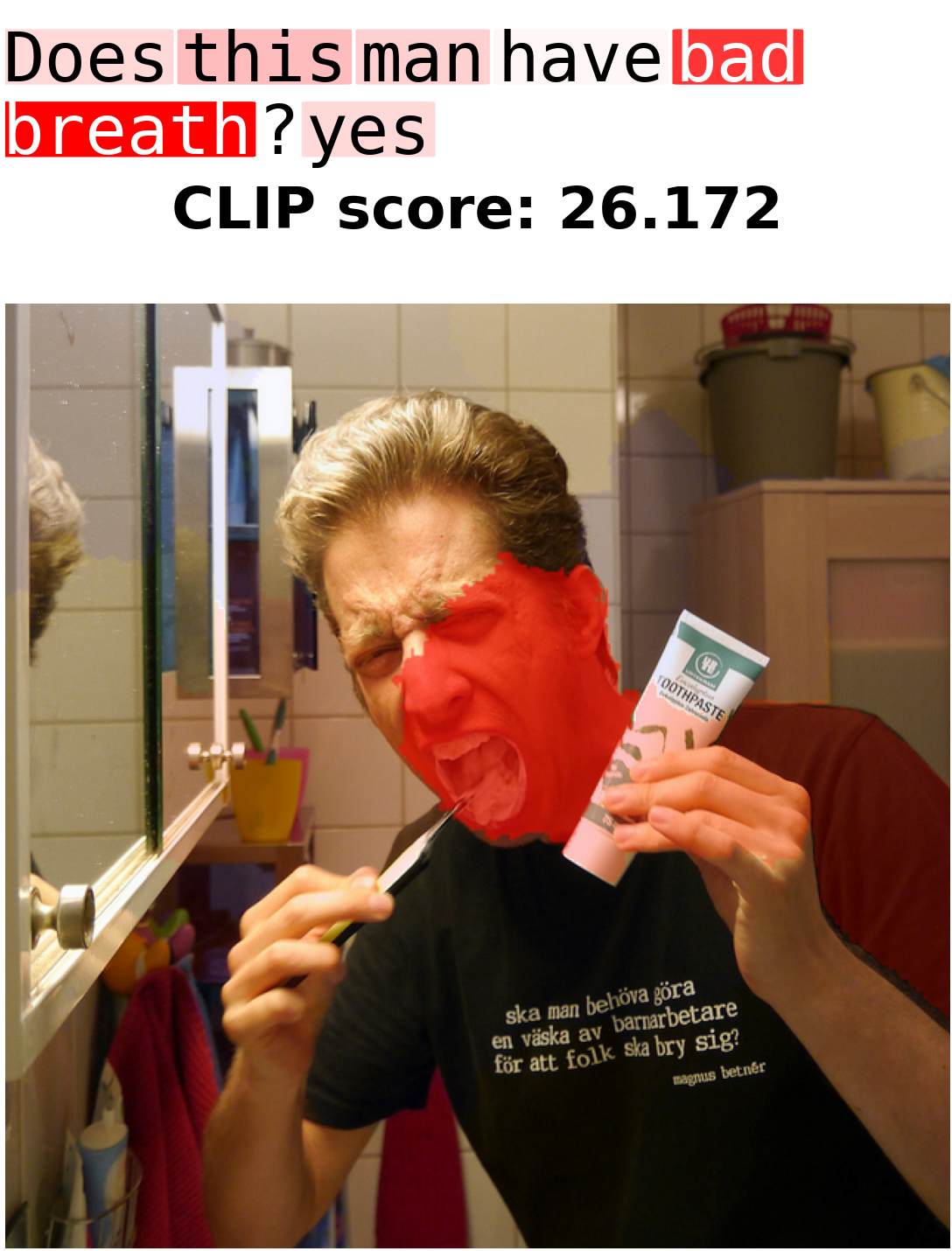}
      \subcaption{DIME}
      \label{fig:clip-dime}
    \end{subfigure}\hfill
    \begin{subfigure}[b]{0.32\linewidth}
      \centering
      \includegraphics[width=\linewidth]{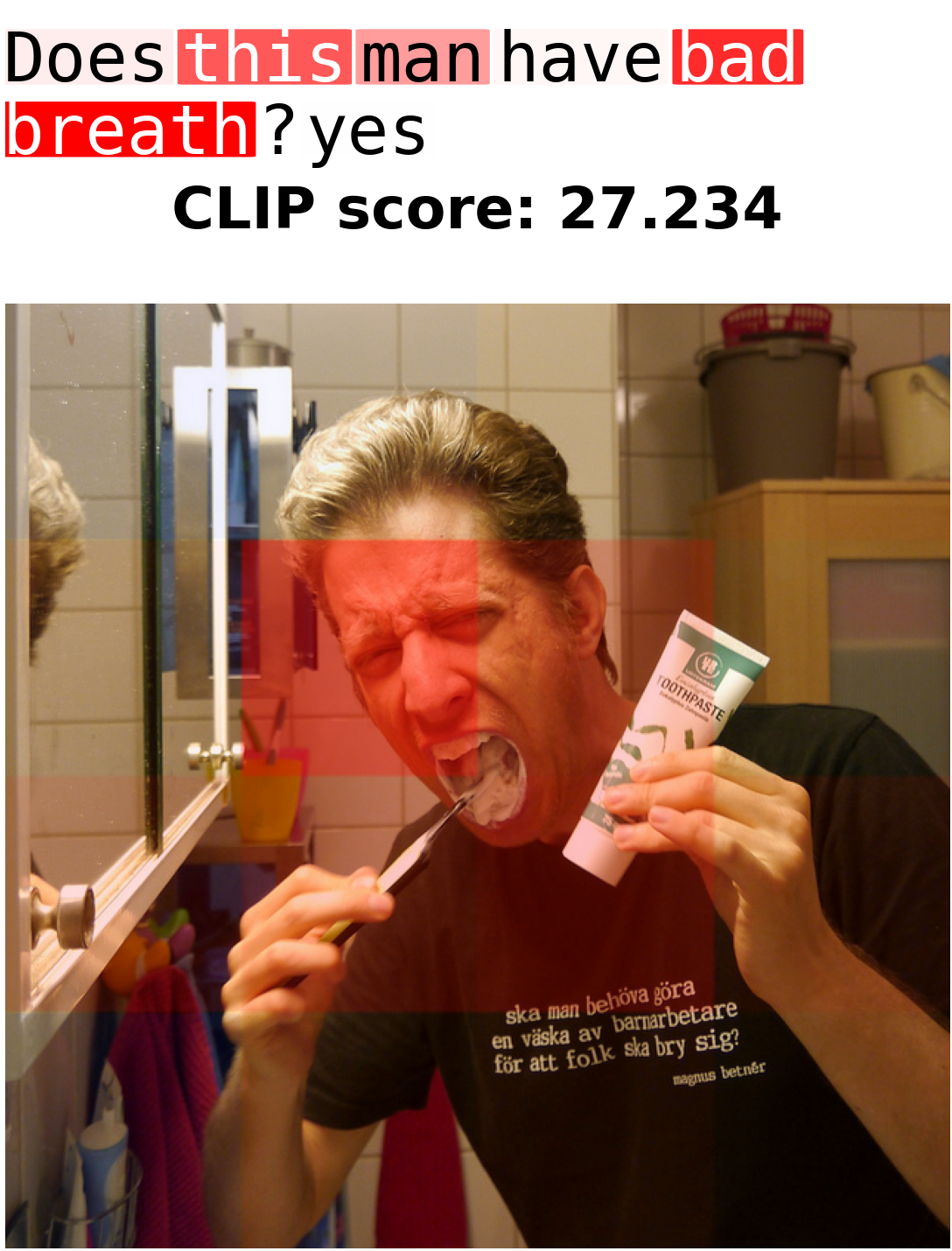}
      \subcaption{MM-SHAP}
      \label{fig:clip-mmshap}
    \end{subfigure}\hfill
    \begin{subfigure}[b]{0.32\linewidth}
      \centering
      \includegraphics[width=\linewidth]{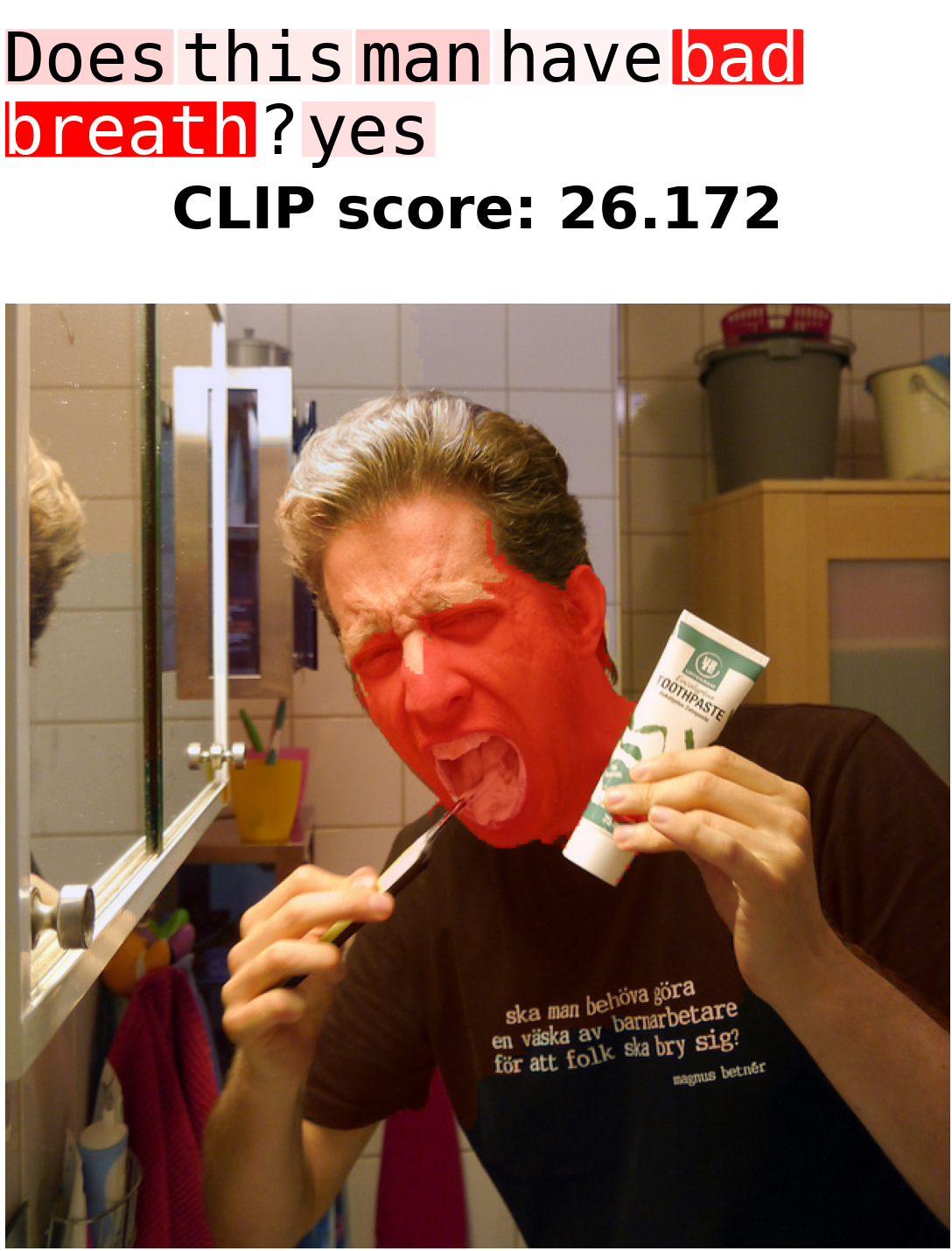}
      \subcaption{LEMON}
      \label{fig:clip-lemon}
    \end{subfigure}
   \end{minipage}

  \noindent
  \begin{minipage}[c]{0.04\linewidth}
    \centering
    {\small\bfseries\rotatebox{90}{LXMERT}}
  \end{minipage}\hfill
  \begin{minipage}[c]{0.95\linewidth}
    \centering
    \begin{subfigure}[b]{0.32\linewidth}
      \centering
      \includegraphics[width=\linewidth]{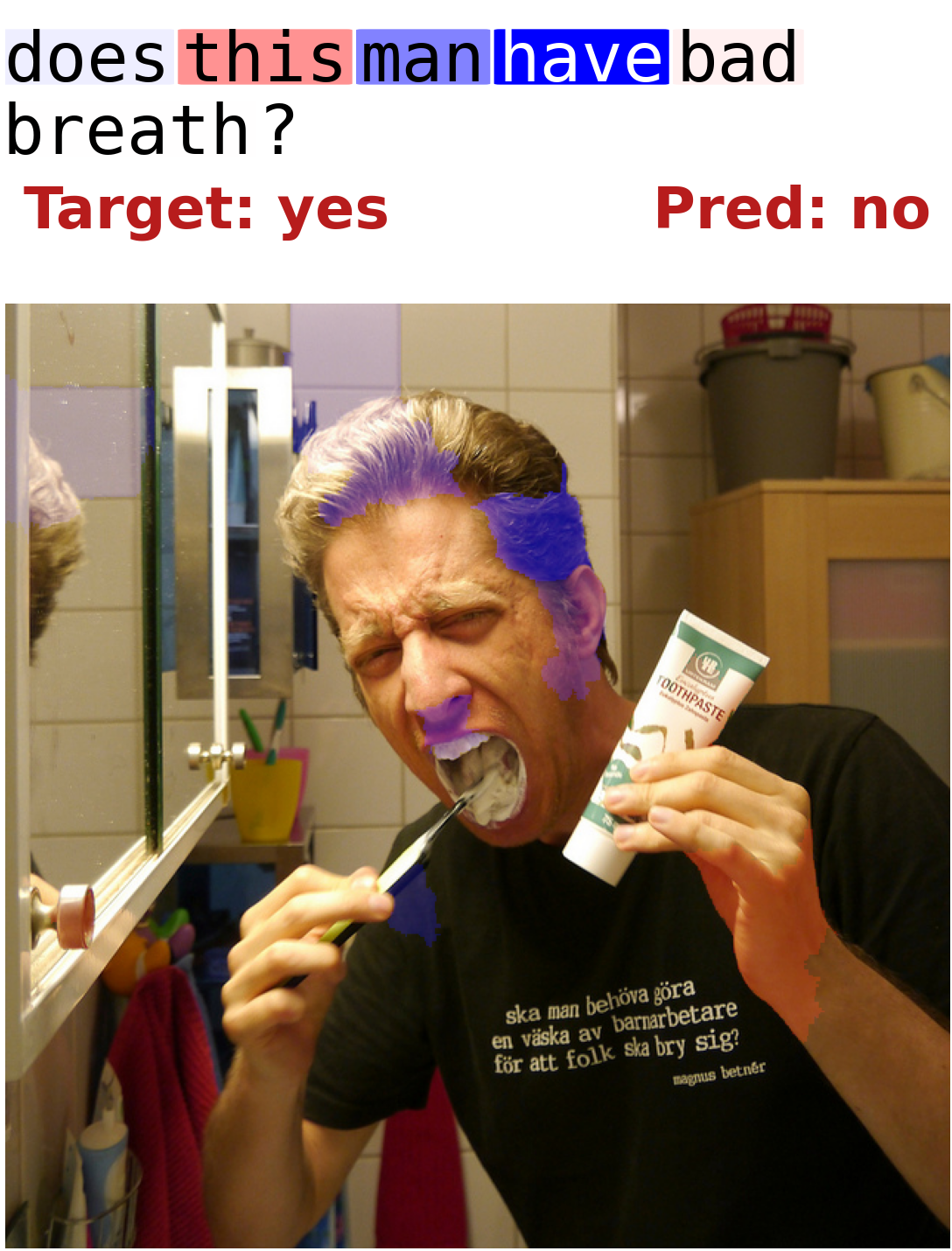}
      \subcaption{DIME}
      \label{fig:lxmert-dime}
    \end{subfigure}\hfill
    \begin{subfigure}[b]{0.32\linewidth}
      \centering
      \includegraphics[width=\linewidth]{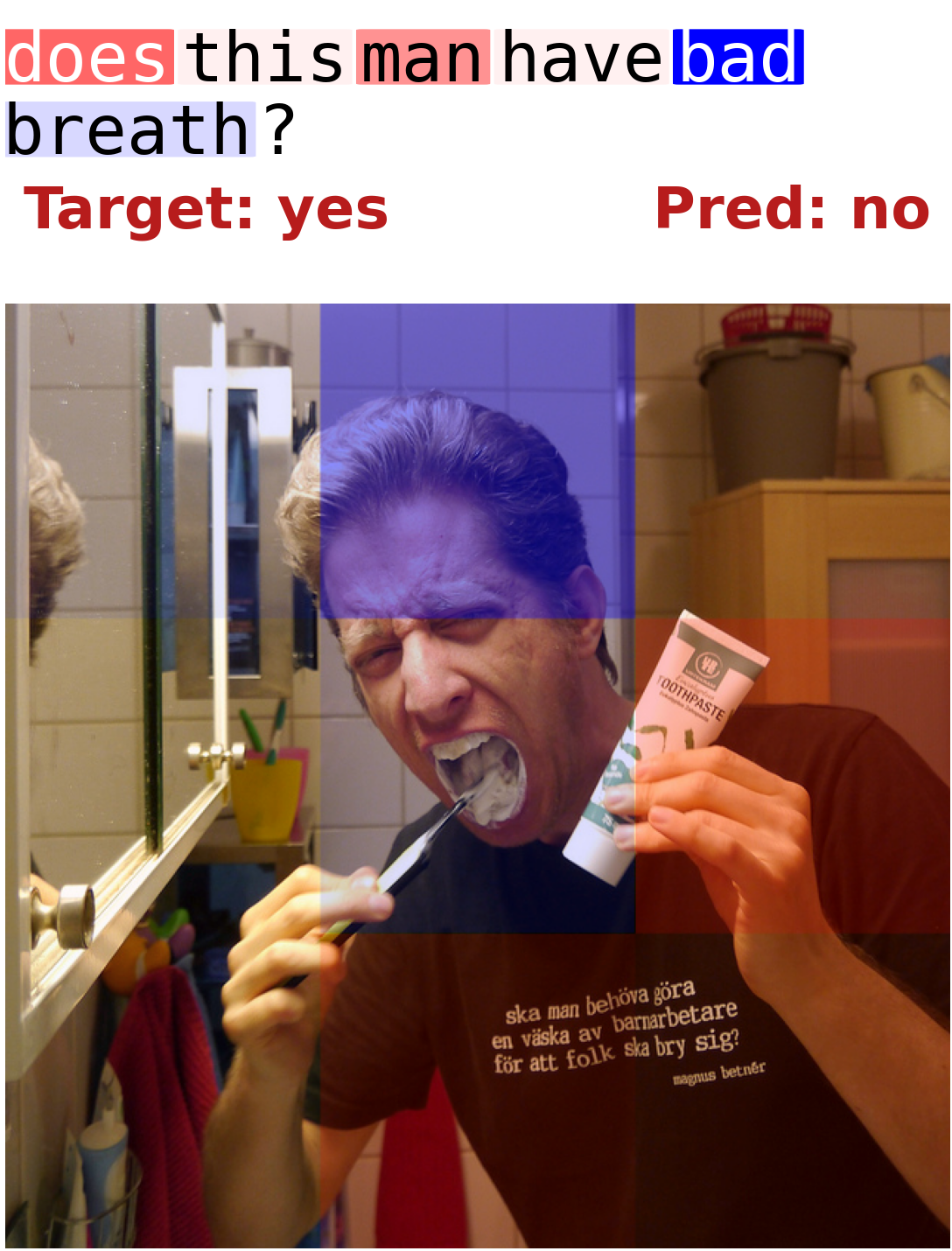}
      \subcaption{MM-SHAP}
      \label{fig:lxmert-mmshap}
    \end{subfigure}\hfill
    \begin{subfigure}[b]{0.32\linewidth}
      \centering
      \includegraphics[width=\linewidth]{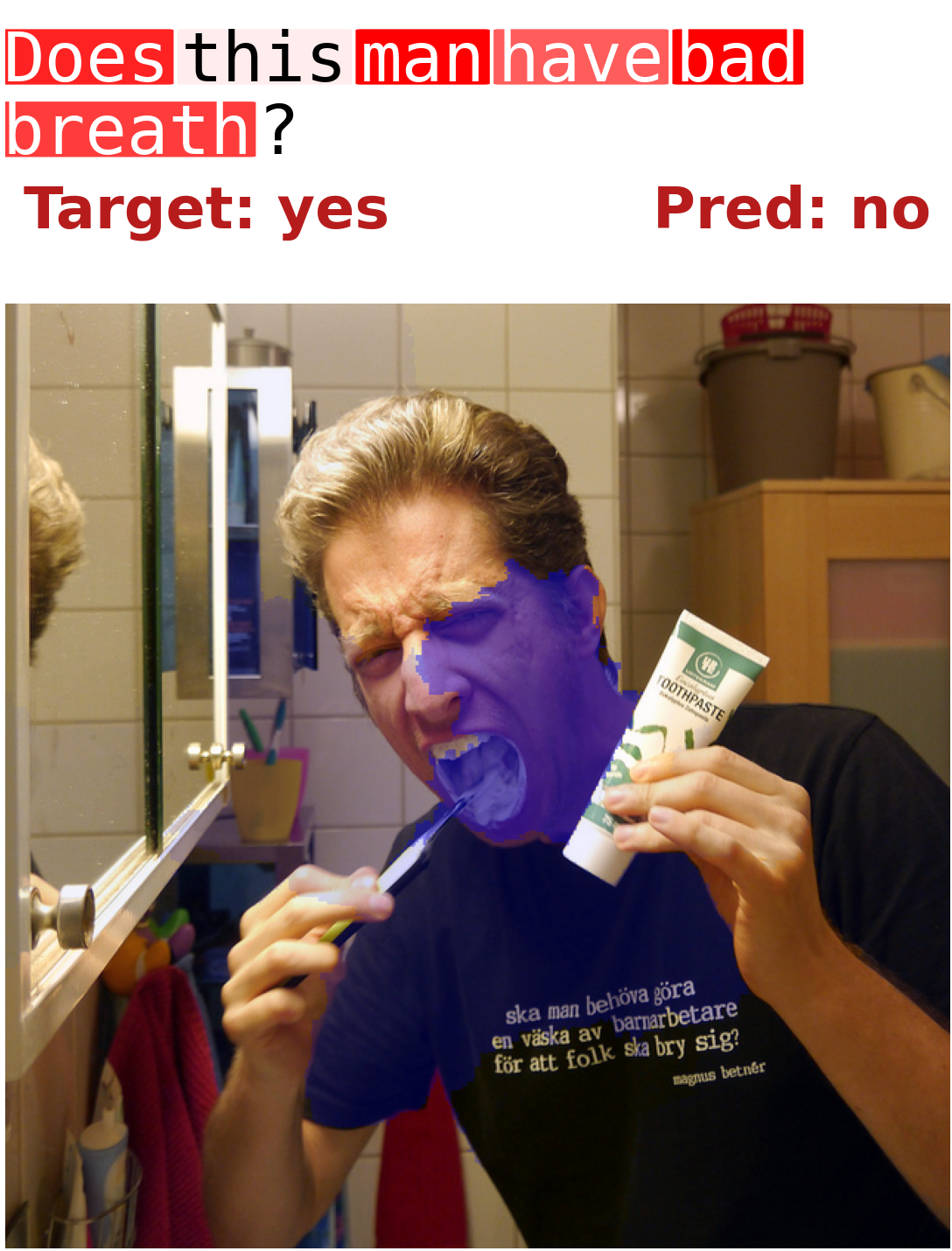}
      \subcaption{LEMON}
      \label{fig:lxmert-lemon}
    \end{subfigure}
   \end{minipage}

  \caption{\textbf{Qualitative comparison across black-boxes and explainers.} Top row (CLIP): CLIP scores the image against the text prompt (question + answer); the higher similarity score indicates strong alignment with the \emph{``yes''} description.
  Bottom row (LXMERT): LXMERT takes (question, image) only and predicts \emph{``no''} (target/GT: \emph{``yes''}); thus most attributions are negative (blue), indicating evidence against the target.
  Across both black-boxes, \textbf{LEMON} consistently attributes the decision to the teeth brushing and mouth area, which is semantically plausible for \emph{bad breath}, while \textbf{MM-SHAP} under-emphasizes the mouth and produces diffuse, coarse evidence and \textbf{DIME} tends to drift towards less relevant hair regions in LXMERT.}
  \label{fig:clip-lxmert-3x2}
\end{figure}

For RQ2, we compare LEMON against two existing model-agnostic multimodal explanation methods: \textbf{MM-SHAP}, which extends Shapley values to multimodal inputs, and \textbf{DIME}, which explains predictions via disentangled multimodal encoders. We evaluate all methods on a multimodal VQA task, considering two different black-box architectures to ensure generality: CLIP and LXMERT, a transformer-based vision–language model. To support a fair and interpretable comparison, we keep the perturbation procedure aligned across methods as far as possible. In particular, for LEMON and DIME we use the same perturbation budget ($N{=}800$) and the same LIME-style image partitioning and masking settings (superpixel segmentation and related perturbation hyperparameters). MM-SHAP is run using the authors’ recommended implementation and settings. We report runtime as the practical cost proxy and evaluate all methods under the same metrics and evaluation protocol.
We additionally report unimodal LIME/SHAP references (image-only / text-only) as context, but we only rank methods within the multimodal block that produces a \emph{single joint} feature ranking. Fig. \ref{fig:clip-lxmert-3x2} qualitatively compares LEMON, DIME, and MM-SHAP for both black-boxes, with LEMON consistently attributing semantically plausible features for \emph{bad breath}, while both MM-SHAP and DIME fail to do so for LXMERT (which mistakenly predicts \emph{no}) and MM-SHAP fails to highlight the mouth/teeth area for CLIP. 

\noindent\textbf{Results on VQA (CLIP).} Table~\ref{tab:vqa_two_backbones} summarizes the results for explaining CLIP on VQA v2 and highlights the trade-off between deletion faithfulness, sparsity, and runtime. 
DIME achieves the strongest deletion faithfulness ($\mathrm{AOPC}_{\mathrm{del}}=1.279\pm2.859$) and stability ($\rho=0.920\pm0.054$), but is significantly more costly ($61.266\pm3.359$\,s; $180.857\pm103.489\times10^3$ forward calls). 
MM-SHAP reduces runtime ($9.823\pm0.744$\,s) but yields lower faithfulness ($0.757\pm2.533$) and larger explanations ($L_0=25.215\pm4.637$).
LEMON provides a better balance: it is fastest by a large margin ($7.511\pm0.888$\,s) and uses much fewer forward calls ($2.714\pm1.559\times10^3$), while remaining close to the top deletion performance ($1.099\pm2.838$ vs.\ DIME $1.279\pm2.859$) with the most compact multimodal explanations ($L_0=20.000\pm4.882$). 
LEMON also maintains high coverage on both modalities (Cov$_\text{img}=0.962\pm0.059$, Cov$_\text{text}=0.996\pm0.017$), and typically ranks first or second on the practical metrics (runtime, query cost, compactness) while keeping faithfulness close to DIME.

\noindent\textbf{Results on VQA (LXMERT).} As for LXMERT, DIME again yields the highest deletion AOPC ($1.412\pm0.652$), but at substantial cost ($42.622\pm2.127$\,s; $263.61\pm151.44\times10^3$ calls) and with much less compact explanations ($L_0=29.017\pm4.345$) and reduced coverage (Cov$_\text{img}=0.772\pm0.076$).
MM-SHAP improves stability ($\rho=0.688\pm0.158$) and yields the highest image coverage (Cov$_\text{img}=0.961\pm0.053$), but it sacrifices faithfulness ($0.991\pm0.483$) and remains relatively costly ($30.676\pm1.642$\,s; $103.622\pm59.532\times10^3$ calls).
LEMON substantially reduces cost while preserving strong faithfulness: it attains near-top deletion performance ($1.373\pm0.392$, second to DIME by $0.039$ in mean AOPC), with the most compact multimodal explanations ($L_0=19.560\pm5.709$) and the lowest runtime among multimodal methods ($20.587\pm1.049$\,s) at markedly lower forward-call budgets ($5.126\pm2.944\times10^3$).
These results suggest LEMON remains effective under explicit multimodal fusion, placing among the top one or two methods across metrics with a better cost--faithfulness--compactness trade-off than multimodal baselines.

\begin{table*}[t]
\centering
\caption{\textbf{VQA v2 results across two black-boxes: \textbf{CLIP} and \textbf{LXMERT}}. We report mean $\pm$std over instances.
\textbf{Multimodal vs unimodal.} (i)/(t) denote image-only and text-only results. The upper block reports \emph{unimodal references} (img-only / text-only), which perturb one modality while keeping the other fixed; the lower block compares \emph{multimodal explainers} that produce a \emph{single joint ranking} over image and text features.
\textbf{Additive costs.} For unimodal methods, the \emph{combined} runtime and forward calls for obtaining both image- and text-side explanations are approximated by summing img-only and text-only costs; we also report the combined explanation size $L_0$ as the sum of img-only and text-only selected features. 
Best results are shown in bold, and second best underlined.} 
\label{tab:vqa_two_backbones}
\small
\setlength{\tabcolsep}{4pt}
\renewcommand{\arraystretch}{1.10}
\resizebox{\textwidth}{!}{%
\begin{tabular}{llrrrrrrr}
\toprule
\textbf{Backbone }& \textbf{Method}
& \textbf{AOPC}$_{\mathrm{del}}\!\uparrow$
& \textbf{Spearman $\rho\!\uparrow$}
& $L_0\!\downarrow$
& \textbf{Cov$_\text{img}\!\uparrow$}
& \textbf{Cov$_\text{text}\!\uparrow$}
& \textbf{Fwd calls ($\times 10^3$)\,$\downarrow$}
& \textbf{Time (s)\,$\downarrow$} \\
\midrule
\multirow{7}{*}{CLIP}
& LIME~\citep{ribeiro2016lime}
& \twolines{$2.046 \pm 2.885\,(\textit{i})$}{$2.795 \pm 2.701\,(\textit{t})$}
& \twolines{$0.985 \pm 0.013\,(\textit{i})$}{$0.953 \pm 0.020\,(\textit{t})$}
& $19.415 \pm 4.389$
& $0.966 \pm 0.044$
& $0.995 \pm 0.020$
& $323.634 \pm 185.930$
& $50.674 \pm 2.341$ \\
& SHAP~\citep{lundberg2017shap}
& \twolines{$0.140 \pm 1.924\,(\textit{i})$}{$4.118 \pm 2.838\,(\textit{t})$}
& \twolines{$0.623 \pm 0.144\,(\textit{i})$}{$0.956 \pm 0.043\,(\textit{t})$}
& $57.695 \pm 2.452$
& $0.540 \pm 0.065$
& $0.995 \pm 0.020$
& $401.100 \pm 162.976$
& $37.317 \pm 0.632$ \\
\cmidrule{2-9}
& DIME~\citep{lyu2022dime}
& \bm{$1.279 \pm 2.859$}
& \bm{$0.920 \pm 0.054$}
& \uline{$21.640 \pm 3.379$}
& \bm{$0.969 \pm 0.042$}
& \bm{$1.000 \pm 0.000$}
& $180.857 \pm 103.489$
& $61.266 \pm 3.359$
 \\
& MM\textnormal{-}SHAP~\citep{parcalabescu2023mmshap}
& $0.757 \pm 2.533$
& \uline{$0.716 \pm 0.117$}
& $25.215 \pm 4.637$
& $0.896 \pm 0.059$
& $0.991 \pm 0.024$
& \uline{$96.030 \pm 55.043$}
& \uline{$9.823 \pm 0.744$} \\
& \textbf{LEMON}
& \uline{$1.099 \pm 2.838$}
& $0.679 \pm 0.141$
& \bm{$20.000 \pm 4.882$}
& \uline{$0.962 \pm 0.059$}
& \uline{$0.996 \pm 0.017$}
& \bm{$2.714 \pm 1.559$}
& \bm{$7.511 \pm 0.888$} \\
\midrule
\multirow{7}{*}{LXMERT}
& LIME~\citep{ribeiro2016lime}
& \twolines{$0.309 \pm 0.392\,(\textit{i})$}{$0.322 \pm 0.303\,(\textit{t})$}
& \twolines{$0.706 \pm 0.174\,(\textit{i})$}{$0.920 \pm 0.098\,(\textit{t})$}
& $25.390 \pm 5.717$
& $0.867 \pm 0.075$
& $1.000 \pm 0.000$
& $3.447 \pm 0.039$
& $98.513 \pm 2.621$ \\

& SHAP~\citep{lundberg2017shap}
& \twolines{$0.305 \pm 0.357\,(\textit{i})$}{$0.445 \pm 0.327\,(\textit{t})$}
& \twolines{$0.155 \pm 0.133\,(\textit{i})$}{$0.726 \pm 0.233\,(\textit{t})$}
& $71.490 \pm 2.505$
& $0.523 \pm 0.119$
& $0.997 \pm 0.013$
& $1.798 \pm 0.010$
& $24.044 \pm 1.321$ \\
\cmidrule{2-9}
& DIME~\citep{lyu2022dime}
& \bm{$1.412 \pm 0.652$}
& $0.554 \pm 0.121$
& $29.017 \pm 4.345$
& $0.772 \pm 0.076$
& \bm{$0.999 \pm 0.005$}
& $263.61 \pm 151.44$
& $42.622 \pm 2.127$ \\
& MM\textnormal{-}SHAP~\citep{parcalabescu2023mmshap}
& $0.991 \pm 0.483$
& \bm{$0.688 \pm 0.158$}
& \uline{$25.515 \pm 4.501$}
& \bm{$0.961 \pm 0.053$}
& $0.998 \pm 0.017$
& \uline{$103.622 \pm 59.532$}
& \uline{$30.676\pm 1.642$} \\
& \textbf{LEMON}
& \uline{$1.373 \pm 0.392$}
& \uline{$0.602 \pm 0.159$}
& \bm{$19.560 \pm 5.709$}
& \uline{$0.871 \pm 0.079$}
& \uline{$0.998 \pm 0.011$}
& \bm{$5.126 \pm 2.944$}
& \bm{$20.587 \pm 1.049$} \\
\bottomrule
\end{tabular}%
}
\end{table*}

\subsection{RQ3: Explanations with CaMCheX \& {REFLACX}}
Finally, RQ3 evaluates LEMON on a real clinical \emph{tri-modal} task, asking whether its highlighted evidence is consistent with available ground-truth annotations while still providing a unified explanation across modalities. We use the recently proposed \textbf{CaMCheX} model for abnormality classification in chest X-rays. CaMCheX is a multimodal transformer-based model that processes a patient’s chest X-ray study along with associated clinical information. We use a tri-modal CaMCheX variant trained with frontal CXR, clinical indications, and vitals. Since REFLACX provides ellipse annotations on the frontal view, evidence-alignment evaluation in RQ3 is assessed on the image modality, while LEMON continues to decompose the same prediction into modality-level contributions and within-modality attributions for \emph{all three} modalities.

We run LEMON on all study-level REFLACX predictions and evaluate image evidence against the corresponding REFLACX ellipse annotations when available. Quantitative evaluation is computed on the subset of predictions for which the required ground-truth evidence and evaluation prerequisites are satisfied; the exact inclusion criteria and implementation details are deferred to Appendix~\ref{app:reflacx_selection}.


RQ3 primarily focuses on \emph{evidence alignment}. We compare LEMON's \emph{positive image evidence} metrics against REFLACX ellipses on the frontal CXR. We report complementary localization summaries including (i) pixel-level ranking quality (PixelAP$^+$), (ii) percentile-threshold IoU summaries (IoU-AUC$^+$), (iii) coverage/precision decompositions at top-$K$ positive regions (GT-Cov@K and MassPrec@K), and (iv) a binary hit criterion based on SPG@K with overlap threshold $\tau$. In parallel, to capture LEMON’s \emph{multimodal} capability, we report \emph{modality contributions} as the modality share of the surrogate weights (image/text/vitals), and visualize modality-wise attributions (image/text/vitals) alongside the combined explanation. Full quantitative results are provided in Table~\ref{ref:reflacx_metrics}.

\begin{figure}[t]
  \centering
  \includegraphics[width=0.90\columnwidth]{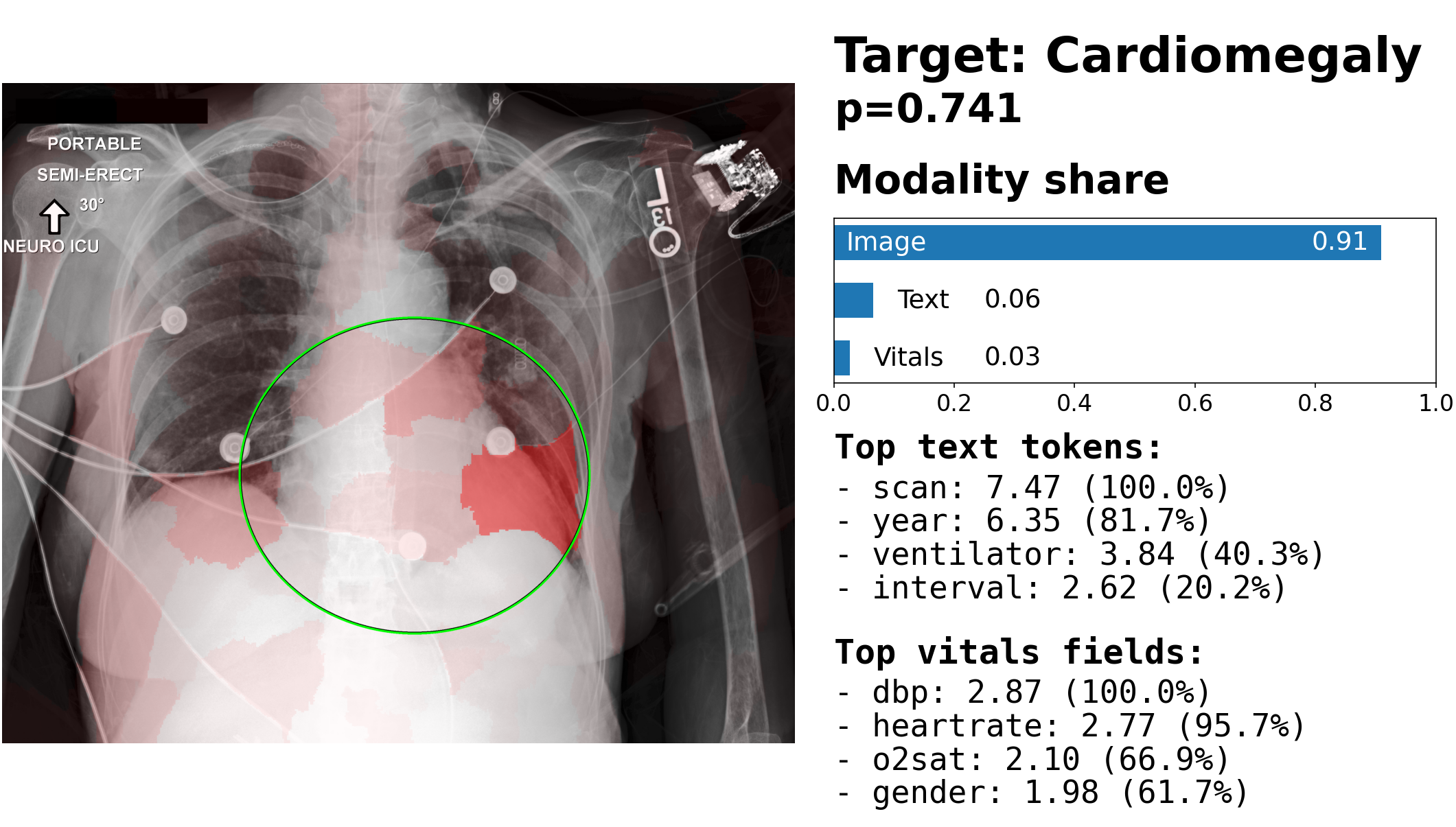}
  \caption{\textbf{Modality-aware clinical explanation on CaMCheX with REFLACX evidence (Cardiomegaly).}
  The card combines: positive image evidence (red heatmap) overlaid on the frontal CXR with the REFLACX ellipse (green), modality contribution via surrogate weight share (image/text/vitals), and the top contributing text tokens and vitals fields. An additional text-dominant example is provided in Appendix~\ref{app:rq3}.}
  \label{fig:reflacx-examples}
\end{figure}

Fig.~\ref{fig:reflacx-examples} provides a LEMON report for a CaMCheX prediction with REFLACX evidence for cardiomegaly, which is often assessed on frontal CXRs via the cardiothoracic ratio (CTR), i.e., the maximal transverse cardiac diameter relative to the thoracic diameter, where values above $\sim$0.5 on a PA view are typically considered abnormal~\citep{truszkiewicz2021ctr}. In our example, LEMON’s positive image evidence concentrates around the enlarged cardiac silhouette and overlaps the radiologist ellipse, providing an intuitive match between the explanation and the annotated evidence.

\begin{table*}[t]
\centering
\caption{\textbf{REFLACX ellipse localization metrics on CaMCheX with LEMON.} We report mean modality shares (image/text/vitals). For other metrics we report median [25th, 75th percentile]. Cells with a light gray background indicate metrics that meet pre-defined heuristic reference thresholds (CH z $\ge 2.0$, PixelAP$^+ \ge 0.45$, IoU-AUC$^+ \ge 0.20$, GT-Cov@K $\ge 95\%$, MassPrec@K $\ge 40\%$, SPG hit $\ge 95\%$). We omit classes with $n<5$ (e.g., Edema) from the by-label panel to avoid unstable estimates. Here $K=8$ and $\tau=0.1$ follow the experimental configuration.}
\label{ref:reflacx_metrics}
\resizebox{\textwidth}{!}{%
\begin{tabular}{lcccccccc}
\toprule
\textbf{Group} & \textbf{$n$} & \textbf{Share (I/T/V) $\uparrow$} & \textbf{CH-$z$ $\uparrow$} & \textbf{PixelAP$^+$ $\uparrow$} & \textbf{IoU-AUC$^+$ $\uparrow$} & \textbf{GT-Cov@K (\%) $\uparrow$} & \textbf{MassPrec@K (\%) $\uparrow$ }& \textbf{SPG hit@$\tau$ (\%) $\uparrow$} \\
\midrule
\multicolumn{9}{l}{\textit{Overall}}\\
ALL
& 24
& 0.68 / 0.26 / 0.06
& \hitcell{2.289} [1.413, 2.450]
& \hitcell{0.497} [0.200, 0.555]
& \hitcell{0.241} [0.117, 0.278]
& \hitcell{97.2} [87.3, 100.0]
& \hitcell{41.2} [23.3, 55.4]
& \hitcell{95.8}
\\
\midrule
\multicolumn{9}{l}{\textit{By label (only $n \ge 5$)}}\\
Atelectasis
& 5
& 0.63 / 0.31 / 0.06
& 1.212 [1.063, 1.481]
& 0.200 [0.195, 0.491]
& 0.128 [0.077, 0.273]
& \hitcell{100.0} [98.3, 100.0]
& 23.9 [23.7, 49.2]
& \hitcell{100.0}
\\
Cardiomegaly
& 11
& 0.75 / 0.19 / 0.06
& \hitcell{2.476} [2.328, 2.572]
& \hitcell{0.504} [0.471, 0.534]
& \hitcell{0.270} [0.262, 0.371]
& 90.4 [88.9, 97.8]
& \hitcell{43.4} [32.3, 50.7]
& \hitcell{100.0}
\\
Pleural Effusion
& 5
& 0.83 / 0.12 / 0.05
& \hitcell{2.322} [2.261, 2.348]
& \hitcell{0.713} [0.617, 0.713]
& \hitcell{0.214} [0.200, 0.226]
& \hitcell{100.0} [98.9, 100.0]
& \hitcell{62.6} [59.2, 63.0]
& \hitcell{100.0}
\\
\bottomrule
\end{tabular}%
}
\end{table*}

On the evaluated REFLACX cases, LEMON’s positive evidence shows consistent alignment with radiologist ellipses under multiple complementary localization criteria (pixel-level ranking quality, percentile-IoU summaries, and top-$K$ coverage/precision decompositions). In particular, the explanations tend to cover a large fraction of the annotated evidence while remaining reasonably concentrated within the ellipses, and the SPG-based hit criterion is frequently satisfied. LEMON also provides a decomposition of modality contributions via the surrogate weight shares, enabling inspection of how much the prediction relies on image, text, and vitals. We report the full quantitative breakdown in Table~\ref{ref:reflacx_metrics}, and include modality-wise visualizations in Fig.~\ref{fig:reflacx-examples}.

\section{Discussions}
\label{sec:discuss}
Across heterogeneous multimodal black-boxes, results indicate that LEMON offers a practical balance between explanation quality and cost. In VQA, LEMON achieves deletion-based faithfulness close to the strongest multimodal baseline, but substantially reduces black-box queries and runtime for both CLIP and LXMERT, suggesting that a single modality-aware surrogate can generalize across multimodal architectures with different fusion structures. One plausible mechanism behind this trade-off is the explicit modality--feature hierarchy induced by group-structured sparsity. The group penalty selects modalities, while the within-group sparsity concentrates attribution on a small set of units inside selected modalities. Empirically, low $L_0$ combined with strong modality coverage supports the interpretation that efficiency gains are not mainly obtained by collapsing to single-modality explanations.

The clinical study further illustrates the value of unified multimodal attributions in settings where decisions may rely on complementary sources of evidence. Although evidence alignment can only be quantitatively assessed on the image modality due to the availability of REFLACX  annotations, LEMON simultaneously decomposes the same CaMCheX prediction into image-, text-, and vitals-level contributions. Quantitative  metrics show consistent overlap between positive image evidence and radiologist annotations, while qualitative examples support intuitive clinical interpretations (e.g., cardiomegaly evidence concentrating around the cardiac silhouette; text dominance when clinical indications explicitly reference the target finding).

LEMON fits a local weighted linear surrogate, so explanations depend on the choice of interpretable units and perturbation design; stability (Spearman's $\rho$) shows higher run-to-run variability than deletion AOPC in our experiments (Table~\ref{tab:vqa_two_backbones}).
In the clinical study, alignment evaluation is further limited by the availability of high-certainty REFLACX ellipse annotations, yielding a smaller evaluated subset.


\section{Conclusions and Future Work}
\label{sec:futurework}
We introduced \textbf{LEMON} (Local Explanations via Modality-aware OptimizatioN), a model-agnostic framework for \textit{local} explanations of multimodal predictors that fits a \textit{single} modality-aware surrogate with \textit{group-structured sparsity} to produce a unified explanation that disentangles \textit{modality-level contributions} from \textit{feature-level attributions}. Across vision--language and clinical multimodal settings, LEMON yields compact structured explanations with competitive deletion-based faithfulness at substantially reduced evaluation cost, supporting practical auditing of multimodal evidence usage.

Future work will focus on improving robustness and stability, which are mainly affected by perturbation stochasticity and sparse surrogate variability under correlated interpretable units. We will pursue three complementary remedies: (i) an engineering fix by increasing the perturbation budget and aggregating explanations across repeated runs, for example by reporting median modality shares or coefficients, (ii) a modeling fix by stabilising the surrogate via bootstrap aggregation or stability selection to encourage consistent support, and (iii) an explanation-level fix by computing modality shares from group-normalised coefficients before aggregation to reduce sensitivity to within-modality scaling. We will also refine perturbations with stratified, modality-balanced, semantically constrained sampling, and extend evaluation to richer modality configurations and broader tasks.

\section*{Acknowledgements}
\label{ack}
This work was supported by the TORUS Project, which has been funded by the UK Engineering and Physical Sciences Research Council (EPSRC), grant number EP/X036146/1.
\section*{Impact Statement}
\label{sec:impact}

This work introduces a model-agnostic method for interpreting multimodal predictions. Our method is designed to facilitate auditing and debugging by improving the transparency of evidence usage, thereby aiding the identification of spurious correlations and potential sources of harm, including high-stakes domains such as healthcare and finance.
As our experiments rely on publicly available datasets and standard pretrained models, any biases, artifacts, or omissions inherent in these resources may propagate into the behaviors we study and the explanations produced. We emphasize that the resulting explanations do not constitute causal claims and should not be treated as definitive justification for a decision. In addition, explanations may be misused for model reverse-engineering or evasion, or may increase unwarranted trust if presented without appropriate context.
We therefore recommend using our explanations alongside complementary verification tools, for example, robustness analysis and domain expert review, particularly in high-stakes applications.

\bibliographystyle{ICML2026/icml2026}
\bibliography{references}

@misc{tian2024drivevlm,
  title         = {DriveVLM: The Convergence of Autonomous Driving and Large Vision-Language Models},
  author        = {Tian, Xiaoyu and Gu, Junru and Li, Bailin and Liu, Yicheng and Wang, Yang and Zhao, Zhiyong and Zhan, Kun and Jia, Peng and Lang, Xianpeng and Zhao, Hang},
  year          = {2024},
  eprint        = {2402.12289},
  archivePrefix = {arXiv},
  primaryClass  = {cs.CV},
  doi           = {10.48550/arXiv.2402.12289}
}

@article{cerneviciene2024xai_finance,
  title     = {Explainable artificial intelligence (XAI) in finance: a systematic literature review},
  author    = {{\v{C}}ernevi{\v{c}}ien\.{e}, Jurgita and Kaba{\v{s}}inskas, Audrius},
  journal   = {Artificial Intelligence Review},
  volume    = {57},
  number    = {216},
  year      = {2024},
  doi       = {10.1007/s10462-024-10854-8}
}

@techreport{wilson2025xai_finance,
  title       = {Explainable {AI} in Finance: Addressing the Needs of Diverse Stakeholders},
  author      = {Wilson, Cheryll-Ann},
  institution = {CFA Institute Research and Policy Center},
  year        = {2025},
  month       = {August},
  url         = {https://rpc.cfainstitute.org/research/reports/2025/explainable-ai-in-finance}
}

@article{thirunavukarasu2023medicine,
  title   = {Large language models in medicine},
  author  = {Thirunavukarasu, A. J. and Ting, D. S. J. and Elangovan, K. and others},
  journal = {Nature Medicine},
  volume  = {29},
  number  = {8},
  pages   = {1930-1940},
  year    = {2023},
  month   = {Aug},
  publisher={Nature Publishing Group},
  doi     = {10.1038/s41591-023-02448-8},
  url     = {https://www.nature.com/articles/s41591-023-02448-8}
}

@article{klein2024finance,
author = {Klein, Tony and Walther, Thomas},
year = {2024},
month = {Nov},
pages = {106358},
title = {Advances in explainable artificial intelligence (xAI)},
volume = {70},
journal = {Finance Research Letters},
doi = {10.1016/j.frl.2024.106358}
}

@article{rosenbacke2024xai_trust,
  title   = {How Explainable Artificial Intelligence Can Increase or Decrease Clinicians’ Trust in AI Applications in Health Care: Systematic Review},
  author  = {Rosenbacke, R. and Melhus, {\AA}. and McKee, M. and Stuckler, D.},
  journal = {JMIR AI},
  volume  = {3},
  pages   = {e53207},
  year    = {2024},
  publisher={JMIR Publications},
  doi     = {10.2196/53207}
}

@article{judijanto2025trust,
  author = {Judijanto, Loso and Achmady, Sayed and Fadhilah, St and Vandika, Arnes},
  year = {2025},
  month = {Apr},
  pages = {344-351},
  title = {The Impact of Explainable AI on User Trust and Ethical AI Adoption in Indonesia},
  volume = {2},
  journal = {The Eastasouth Journal of Information System and Computer Science},
  doi = {10.58812/esiscs.v2i03.531}
}

@article{ma2024ftd_differential,
  title   = {Differential diagnosis of frontotemporal dementia subtypes with explainable deep learning on structural MRI},
  author  = {Ma, D. and Stocks, J. and Rosen, H. J. and Kantarci, K. and Lockhart, S. N. and Bateman, J. R. and Craft, S. and Gurcan, M. N. and Popuri, K. and Beg, M. F. and Wang, L.},
  journal = {Frontiers in Neuroscience},
  volume  = {18},
  pages   = {1331677},
  year    = {2024},
  month   = {Feb},
  doi     = {10.3389/fnins.2024.1331677}
}

@inproceedings{radford2021clip,
  title={Learning Transferable Visual Models From Natural Language Supervision}, 
  author={Alec Radford and Jong Wook Kim and Chris Hallacy and Aditya Ramesh and Gabriel Goh and Sandhini Agarwal and Girish Sastry and Amanda Askell and Pamela Mishkin and Jack Clark and Gretchen Krueger and Ilya Sutskever},
  year={2021},
  eprint={2103.00020},
  archivePrefix={arXiv},
  primaryClass={cs.CV},
  url={https://arxiv.org/abs/2103.00020}
}

@misc{liu2023llava,
  title={Visual Instruction Tuning}, 
  author={Haotian Liu and Chunyuan Li and Qingyang Wu and Yong Jae Lee},
  year={2023},
  eprint={2304.08485},
  archivePrefix={arXiv},
  primaryClass={cs.CV},
  url={https://arxiv.org/abs/2304.08485}, 
}

@misc{dang2024gap,
  title={Explainable and Interpretable Multimodal Large Language Models: A Comprehensive Survey}, 
  author={Yunkai Dang and Kaichen Huang and Jiahao Huo and Yibo Yan and Sirui Huang and Dongrui Liu and Mengxi Gao and Jie Zhang and Chen Qian and Kun Wang and Yong Liu and Jing Shao and Hui Xiong and Xuming Hu},
  year={2024},
  eprint={2412.02104},
  archivePrefix={arXiv},
  primaryClass={cs.CL},
  url={https://arxiv.org/abs/2412.02104}, 
}

@misc{agarwal2025rethink,
  title={Rethinking Explainability in the Era of Multimodal AI}, 
  author={Chirag Agarwal},
  year={2025},
  eprint={2506.13060},
  archivePrefix={arXiv},
  primaryClass={cs.AI},
  url={https://arxiv.org/abs/2506.13060}, 
}

@misc{jethani2022fastshap,
  title={FastSHAP: Real-Time Shapley Value Estimation}, 
  author={Neil Jethani and Mukund Sudarshan and Ian Covert and Su-In Lee and Rajesh Ranganath},
  year={2022},
  eprint={2107.07436},
  archivePrefix={arXiv},
  primaryClass={stat.ML},
  url={https://arxiv.org/abs/2107.07436}, 
}

@misc{kelodjou2024shaping,
  title={Shaping Up SHAP: Enhancing Stability through Layer-Wise Neighbor Selection}, 
  author={Gwladys Kelodjou and Laurence Rozé and Véronique Masson and Luis Galárraga and Romaric Gaudel and Maurice Tchuente and Alexandre Termier},
  year={2024},
  eprint={2312.12115},
  archivePrefix={arXiv},
  primaryClass={cs.LG},
  url={https://arxiv.org/abs/2312.12115}, 
}

@inproceedings{parcalabescu2023mmshap,
  title     = {MM-SHAP: A Performance-Agnostic Metric for Measuring Multimodal Contributions in Vision and Language Models},
  author    = {Parcalabescu, Letitia and Frank, Anette},
  booktitle = {Proceedings of the 61st Annual Meeting of the Association for Computational Linguistics (ACL)},
  year      = {2023}
}

@misc{ribeiro2016lime,
  title={"Why Should I Trust You?": Explaining the Predictions of Any Classifier}, 
  author={Marco Tulio Ribeiro and Sameer Singh and Carlos Guestrin},
  year={2016},
  eprint={1602.04938},
  archivePrefix={arXiv},
  primaryClass={cs.LG},
  url={https://arxiv.org/abs/1602.04938}, 
}

@article{simon2013sparsegrouplasso,
  author = {Simon, Noah and Friedman, Jerome and Hastie, Trevor and Tibshirani, Robert},
  year = {2013},
  month = {Apr},
  title = {A Sparse-Group Lasso},
  volume = {22},
  journal = {Journal of Computational and Graphical Statistics},
  doi = {10.1080/10618600.2012.681250}
}

@article{zhao2023survey,
  title = "Explainability for Large Language Models: A Survey",
  author = "Haiyan Zhao and Hanjie Chen and Fan Yang and Ninghao Liu and Huiqi Deng and Hengyi Cai and Shuaiqiang Wang and Dawei Yin and Mengnan Du",
  year = "2024",
  month = Feb,
  day = "22",
  journal = "ACM Transactions on Intelligent Systems and Technology",
  publisher = "Association for Computing Machinery (ACM)",
  volume = {15},
  number = {2},
  issn = {2157-6904},
}

@inproceedings{selvaraju2017gradcam,
  title     = {Grad-CAM: Visual Explanations from Deep Networks via Gradient-Based Localization},
  author    = {Selvaraju, Ramprasaath R. and Cogswell, Michael and Das, Abhishek and others},
  booktitle = {Proceedings of the IEEE International Conference on Computer Vision},
  year      = {2017}
}

@misc{lu2019vilbert,
  title={ViLBERT: Pretraining Task-Agnostic Visiolinguistic Representations for Vision-and-Language Tasks}, 
  author={Jiasen Lu and Dhruv Batra and Devi Parikh and Stefan Lee},
  year={2019},
  eprint={1908.02265},
  archivePrefix={arXiv},
  primaryClass={cs.CV},
  url={https://arxiv.org/abs/1908.02265}, 
}

@InProceedings{chefer2021generic,
    author    = {Chefer, Hila and Gur, Shir and Wolf, Lior},
    title     = {Transformer Interpretability Beyond Attention Visualization},
    booktitle = {Proceedings of the IEEE/CVF Conference on Computer Vision and Pattern Recognition (CVPR)},
    month     = {June},
    year      = {2021},
    pages     = {782-791}
}

@misc{tan2019lxmert,
  title={LXMERT: Learning Cross-Modality Encoder Representations from Transformers}, 
  author={Hao Tan and Mohit Bansal},
  year={2019},
  eprint={1908.07490},
  archivePrefix={arXiv},
  primaryClass={cs.CL},
  url={https://arxiv.org/abs/1908.07490}, 
}

@misc{lopardo2024attention,
  title={Attention Meets Post-hoc Interpretability: A Mathematical Perspective}, 
  author={Gianluigi Lopardo and Frederic Precioso and Damien Garreau},
  year={2024},
  eprint={2402.03485},
  archivePrefix={arXiv},
  primaryClass={stat.ML},
  url={https://arxiv.org/abs/2402.03485}, 
}

@misc{lyu2024nlp,
  title={Towards Faithful Model Explanation in NLP: A Survey}, 
  author={Qing Lyu and Marianna Apidianaki and Chris Callison-Burch},
  year={2024},
  eprint={2209.11326},
  archivePrefix={arXiv},
  primaryClass={cs.CL},
  url={https://arxiv.org/abs/2209.11326}, 
}

@misc{lyu2022dime,
  title={DIME: Fine-grained Interpretations of Multimodal Models via Disentangled Local Explanations}, 
  author={Yiwei Lyu and Paul Pu Liang and Zihao Deng and Ruslan Salakhutdinov and Louis-Philippe Morency},
  year={2022},
  eprint={2203.02013},
  archivePrefix={arXiv},
  primaryClass={cs.LG},
  url={https://arxiv.org/abs/2203.02013}, 
}

@misc{sotirou2025musiclime,
  title={MusicLIME: Explainable Multimodal Music Understanding}, 
  author={Theodoros Sotirou and Vassilis Lyberatos and Orfeas Menis Mastromichalakis and Giorgos Stamou},
  year={2025},
  eprint={2409.10496},
  archivePrefix={arXiv},
  primaryClass={cs.SD},
  url={https://arxiv.org/abs/2409.10496}, 
}

@misc{lundberg2017shap,
  title={A Unified Approach to Interpreting Model Predictions}, 
  author={Scott Lundberg and Su-In Lee},
  year={2017},
  eprint={1705.07874},
  archivePrefix={arXiv},
  primaryClass={cs.AI},
  url={https://arxiv.org/abs/1705.07874}, 
}

@article{Ranjbaran2025cshap,
  author = {Ranjbaran, Golshid and Reforgiato Recupero, Diego and Roy, Chanchal and Schneider, Kevin},
  year = {2025},
  month = {01},
  title = {C-SHAP: A Hybrid Method for Fast and Efficient Interpretability},
  volume = {15},
  journal = {Applied Sciences},
  doi = {10.3390/app15020672}
}

@misc{wang2025multishap,
  title={MultiSHAP: A Shapley-Based Framework for Explaining Cross-Modal Interactions in Multimodal AI Models}, 
  author={Zhanliang Wang and Kai Wang},
  year={2025},
  eprint={2508.00576},
  archivePrefix={arXiv},
  primaryClass={cs.AI},
  url={https://arxiv.org/abs/2508.00576}, 
}

@article{pennisi2024diffexplainer,
  author = {Pennisi, Matteo and Bellitto, Giovanni and Palazzo, Simone and Kavasidis, Isaak and Shah, Mubarak and Spampinato, Concetto},
  year = {2025},
  month = {Oct},
  pages = {104559},
  title = {DiffExplainer: Towards cross-modal global explanations with diffusion models},
  volume = {262},
  journal = {Computer Vision and Image Understanding},
  doi = {10.1016/j.cviu.2025.104559}
}

@misc{schwab2019cxplain,
  title={CXPlain: Causal Explanations for Model Interpretation under Uncertainty}, 
  author={Patrick Schwab and Walter Karlen},
  year={2019},
  eprint={1910.12336},
  archivePrefix={arXiv},
  primaryClass={cs.LG},
  url={https://arxiv.org/abs/1910.12336}, 
}

@misc{goyal2017vqa2,
  title={Making the V in VQA Matter: Elevating the Role of Image Understanding in Visual Question Answering}, 
  author={Yash Goyal and Tejas Khot and Douglas Summers-Stay and Dhruv Batra and Devi Parikh},
  year={2017},
  eprint={1612.00837},
  archivePrefix={arXiv},
  primaryClass={cs.CV},
  url={https://arxiv.org/abs/1612.00837}, 
}

@article{Bigolin2022reflacx,
   title={{REFLACX}, a dataset of reports and eye-tracking data for localization of abnormalities in chest x-rays},
   volume={9},
   ISSN={2052-4463},
   url={http://dx.doi.org/10.1038/s41597-022-01441-z},
   DOI={10.1038/s41597-022-01441-z},
   number={1},
   journal={Scientific Data},
   publisher={Springer Science and Business Media LLC},
   author={Bigolin Lanfredi, Ricardo and Zhang, Mingyuan and Auffermann, William F. and Chan, Jessica and Duong, Phuong-Anh T. and Srikumar, Vivek and Drew, Trafton and Schroeder, Joyce D. and Tasdizen, Tolga},
   year={2022},
   month={june}
}

@misc{sloan2025camchex,
  title={Clinically-aligned Multi-modal Chest X-ray Classification}, 
  author={Phillip Sloan and Edwin Simpson and Majid Mirmehdi},
  year={2025},
  eprint={2511.09581},
  archivePrefix={arXiv},
  primaryClass={eess.IV},
  url={https://arxiv.org/abs/2511.09581}, 
}

@article{truszkiewicz2021ctr,
  title   = {Radiological Cardiothoracic Ratio in Evidence-Based Medicine},
  author  = {Truszkiewicz, Krystian and Por{\k{e}}ba, Rafa{\l} and Ga{\'c}, Pawe{\l}},
  journal = {Journal of Clinical Medicine},
  year    = {2021},
  month   = {May},
  volume  = {10},
  number  = {9},
  pages   = {2016},
  doi     = {10.3390/jcm10092016}
}

\newpage
\appendix
\onecolumn
\appendix
\section{Selecting modality weights $\alpha$.}
\label{app:alpha}
The vector $\alpha=(\alpha_1,\dots,\alpha_M)$ controls how strongly deviations in each modality contribute to locality. We initialise modality weights by setting
\begin{equation}
\alpha_m^{\mathrm{base}}\propto \frac{1}{\operatorname{IQR}_{i\in[N]} d_m(z'_i) + \epsilon_d}, \qquad \sum_m \alpha_m^{\mathrm{base}}=1,
\end{equation}
so that modalities with more concentrated perturbation distances are not overly down-weighted.
Around this base point we define a small logarithmic grid $\mathcal{A}$ and select $\alpha$ on a train/validation split of $\mathcal{D}$.

For each candidate $\alpha\in\mathcal{A}$, we compute $D(z'_i)$ and weights $w_i=w(z'_i)$, fit a local surrogate on the train split, and evaluate on the validation split a scalarised objective that balances fidelity and non-degenerate locality.
Specifically, we define the weighted coefficient of determination
\begin{equation}
\mathrm{WR}^2(\alpha) =
1-\frac{\sum_{i\in\mathcal{V}} w_i\big(y'_i-(\hat\beta_0+z'_i{}^\top\hat\beta)\big)^2}{\sum_{i\in\mathcal{V}} w_i\big(y'_i-\bar y'\big)^2},
\end{equation}
which is the standard $R^2$ generalized to the weighted setting, and the effective number of samples
\begin{equation}
\mathrm{Neff}(\alpha) = \frac{\big(\sum_{i\in\mathcal{V}} w_i\big)^2}{\sum_{i\in\mathcal{V}} w_i^2}.
\end{equation}
We then maximise
\begin{equation}
J(\alpha)=\mathrm{WR}^2(\alpha)
- \lambda_{\mathrm{deg}}\,
\mathbf{1}\!\big[\mathrm{Neff}(\alpha)<N_{\min}\big],
\end{equation}
where $N_{\min}$ is a minimum effective sample size and $\lambda_{\mathrm{deg}}$ penalises degenerate kernels that concentrate weight on too few perturbations.
The selected $\alpha^\star=\arg\max_{\alpha\in\mathcal{A}} J(\alpha)$ is used to recompute $D(z')$ and $w(z')$ on the full neighbourhood before fitting the final surrogate.

\section{Group reweighting.}
\label{app:weights}
For each group $g$ with size $p_g=|g|$, let $\mathrm{sd}_w(z'_g)$ denote the mean weighted standard deviation of the corresponding columns of $z'$. We set
\begin{equation}
\tau_g=\sqrt{p_g}\cdot \gamma_g,\qquad 
\gamma_g\ \propto\ \frac{1}{\overline{\mathrm{sd}}_w(z'_{g})},
\end{equation}
and rescale $\gamma_g$ to have mean~1 across groups.
This offsets both group-size effects and groups whose features are rarely switched on.

As is standard with $\ell_1$-type penalties, the coefficients of active features can be biased towards zero. After solving~\eqref{eq:lemon_obj}, we optionally perform a \emph{support-preserving ridge refit} restricted to the selected support, keeping inactive coefficients fixed at zero, to improve local fidelity without sacrificing sparsity.

\section{Implementation Details and Reproducibility}
\label{app:implementation}
\paragraph{Backbone models and hardware.}
We evaluate LEMON on two representative multimodal black-box predictors: CLIP for vision--language question answering and LXMERT for multimodal fusion tasks, as well as the CaMCheX clinical model for tri-modal prediction.
All experiments are implemented in PyTorch and executed on an HPC cluster using NVIDIA A100 GPUs (Ampere architecture).
Each compute node provides 4 GPUs and approximately 250\,GB host memory under the \texttt{ampere} partition.

\paragraph{Hyperparameter configuration.}
Hyperparameters were selected via grid search separately for each backbone and dataset on held-out validation data.
For VQA with CLIP, we draw $N{=}800$ perturbations and fit a Sparse Group Lasso surrogate with $\alpha{=}0.004$ and $\ell_1$ ratio $0.9$; deletion faithfulness is evaluated with 20 steps.
For VQA with LXMERT, we use $N{=}800$ perturbations (batch size 32), superpixel settings $(k{=}8,\; d_{\max}{=}100,\; r{=}0.2)$, and SGL parameters $\alpha{=}0.02$ and $\ell_1$ ratio $0.9$, again with 20 deletion steps.
For CaMCheX, query-superpixel parameters are set to target side 384, kernel size 8, maximum distance 70, and ratio 0.15; clinical evaluation uses probability threshold $0.7$ and minimum certainty threshold 4.
All methods are evaluated under matched query budgets.

\paragraph{Randomness and stability.}
We do not fix a global random seed.
Instead, robustness is assessed through repeated explanation runs.
For CaMCheX, each case is evaluated over 50 perturbation rolls.
For CLIP-VQA and LXMERT-VQA, repeated runs are performed using the stability mode.
Reported metrics are aggregated over evaluation samples, and stability is quantified using Spearman's rank correlation across repeated runs.

\section{RQ3}
\label{app:rq3}

\subsection{Ground-truth alignment metrics on REFLACX}
For RQ3 we evaluate whether LEMON’s highlighted \emph{positive} image evidence overlaps with radiologist-provided ellipse annotations from REFLACX. Let $H_{\text{pos}}\in[0,1]^{H\times W}$ denote the resized positive-evidence heatmap, constructed by aggregating superpixel masks weighted by the \emph{positive part} of the image surrogate coefficients and normalizing to $[0,1]$. Let $G\in\{0,1\}^{H\times W}$ be the binary ground-truth mask obtained by rasterizing the REFLACX ellipse(s) for the target finding on the frontal image.

\paragraph{Modality weight share.}
To quantify modality contributions, we compute the share of surrogate mass per modality:
\begin{equation}
s_m \;=\; \frac{\sum_{j\in \mathcal{I}_m} |w_j|}{\sum_{j} |w_j|},
\qquad m\in\{\text{image},\text{text},\text{vitals}\},
\end{equation}
where $w_j$ are surrogate coefficients and $\mathcal{I}_m$ indexes features belonging to modality $m$.

\paragraph{Contrast Heat z-score (positive), CH-$z$.}
We first compute the positive contrast heat
\begin{equation}
\mathrm{CH}_{\text{pos}} \;=\; \mathbb{E}\!\left[H_{\text{pos}} \mid G=1\right] \;-\; \mathbb{E}\!\left[H_{\text{pos}} \mid G=0\right].
\end{equation}
To contextualize localization strength, we construct a roll-based baseline by randomly shifting $G$ (preserving shape/area) and recomputing $\mathrm{CH}_{\text{pos}}$ over $R$ rolls. We then report
\begin{equation}
\mathrm{CH}\text{-}z \;=\; \frac{\mathrm{CH}_{\text{pos}} - \mu_{\text{roll}}}{\sigma_{\text{roll}}+\epsilon},
\end{equation}
where $(\mu_{\text{roll}},\sigma_{\text{roll}})$ are the roll mean and std, and $\epsilon$ is a small constant for numerical stability.

\paragraph{Pixel-level ranking quality (positive), PixelAP$^+$.}
We treat each pixel’s heat value $H_{\text{pos}}(u,v)$ as a score and the ground-truth mask $G(u,v)$ as the binary label, and compute the average precision (AP) over all pixels:
\begin{equation}
\mathrm{PixelAP}^+ \;=\; \mathrm{AP}\big(\{H_{\text{pos}}(u,v)\}, \{G(u,v)\}\big).
\end{equation}

\paragraph{Percentile IoU curve and IoU-AUC$^+$.}
We threshold the heatmap by percentiles to obtain a family of binary masks. For percentile $p$, let $t_p$ be the $p$-th percentile of $H_{\text{pos}}$ values (typically computed on non-zero pixels), and define
\begin{equation}
M_p(u,v) \;=\; \mathbb{I}\!\left[H_{\text{pos}}(u,v)\ge t_p\right].
\end{equation}
We compute $\mathrm{IoU}(p)=\frac{|M_p\cap G|}{|M_p\cup G|}$ over a percentile range and summarize via the area under the curve:
\begin{equation}
\mathrm{IoU\text{-}AUC}^+ \;=\; \frac{1}{|\mathcal{P}|}\sum_{p\in\mathcal{P}} \mathrm{IoU}(p),
\end{equation}
where $\mathcal{P}$ is a fixed percentile grid (e.g., $p\in\{90,\dots,99\}$).

\paragraph{Top-$K$ positive superpixel overlap metrics (positive).}
Let $\{S_j\}_{j=1}^{K_{\text{img}}}$ be the image superpixel masks and $w_j$ the corresponding image surrogate weights.
We consider \emph{positive} superpixels with $w_j>0$ and select the top-$K$ indices $\mathcal{K}$ by descending $w_j$.
Define the area $a_j=|S_j|$, overlap $o_j=|S_j\cap G|$, and overlap ratio $r_j=o_j/\max(a_j,1)$.
We report:
\begin{align}
\mathrm{GT\text{-}Cov@K}(\%) \;&=\; 100\cdot \frac{\sum_{j\in\mathcal{K}} o_j}{|G|},\\
\mathrm{MassPrec@K}(\%) \;&=\; 100\cdot \frac{\sum_{j\in\mathcal{K}} w_j\, r_j}{\sum_{j\in\mathcal{K}} w_j},\\
\mathrm{SPG\ hit@}\tau(\%) \;&=\; 100\cdot \mathbb{I}\!\left[\max_{j\in\mathcal{K}} r_j \ge \tau\right],
\end{align}
where $\tau$ is an overlap-ratio threshold (e.g., $\tau=0.1$). For a group of $n$ cases, we report the hit rate as
$\mathrm{SPG\ hit@}\tau(\%) = 100\cdot \frac{1}{n}\sum_{i=1}^{n}\mathbb{I}\!\left[\max_{j\in\mathcal{K}_i} r_{ij} \ge \tau\right]$.

\subsection{Additional setup details and results}
\label{app:reflacx_selection}
We run LEMON across the full REFLACX subset under the tri-modal CaMCheX setting (frontal CXR, clinical text, vitals). Quantitative alignment metrics require a well-defined target finding with available REFLACX ellipses; thus, metrics are computed on the subset of (study, finding) pairs that satisfy the evaluation prerequisites implemented in our pipeline (details below).

\paragraph{Evaluation set construction.}
For each REFLACX study, we consider CaMCheX findings whose predicted probability exceeds a confidence threshold (default $p\ge 0.7$) and retain only targets with at least one corresponding REFLACX ellipse on the frontal view with high radiologist certainty (default certainty $\ge 4$). This results in $n=24$ evaluated (study, finding) pairs in the present run, with the following breakdown: Atelectasis ($n=5$), Cardiomegaly ($n=11$), Pleural Effusion ($n=5$), and Edema ($n=3$).

\begin{figure}[!t]
  \centering
  \includegraphics[width=\columnwidth]{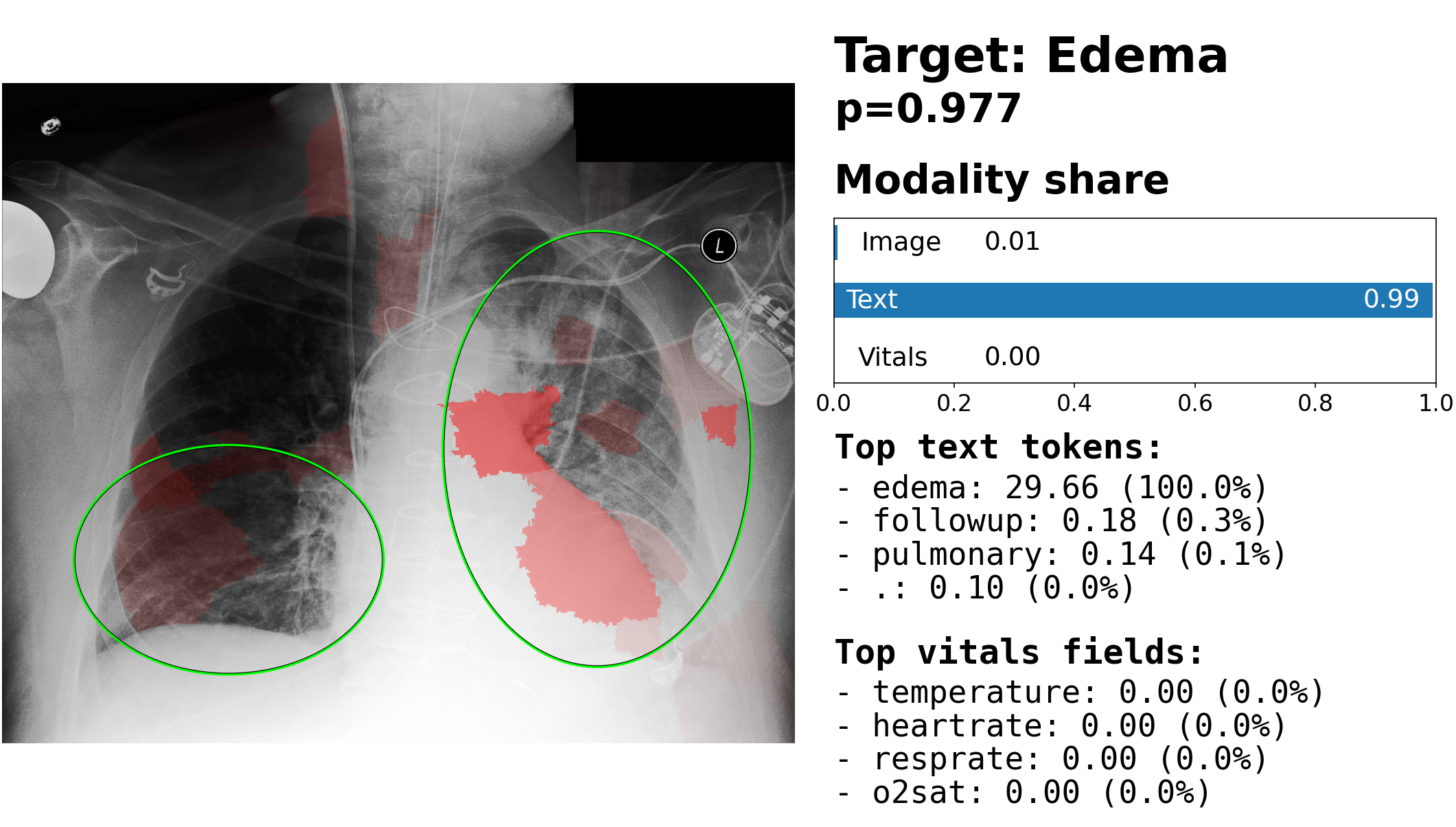}
  \caption{\textbf{Additional modality-aware explanation (Edema).}
  Same layout as Fig.~\ref{fig:reflacx-examples}.}
  \label{fig:reflacx-examples-edema}
\end{figure}

\paragraph{Summary metrics.}
Table~\ref{ref:reflacx_metrics} reports modality weight shares (mean) and evidence-alignment metrics (median [25th, 75th]) including CH-$z$, PixelAP$^+$, IoU-AUC$^+$, GT-Cov@K, MassPrec@K, and SPG hit@$\tau$. For per-finding summaries, we report results only for findings with $n\ge 5$ to avoid unstable estimates in small strata.

\paragraph{Text-dominant case.}
Figure~\ref{fig:reflacx-examples-edema} illustrates a case where the text modality dominates the local explanation: the clinical indications explicitly mention edema-related findings, so the surrogate assigns most positive mass to text units, while image and tabular evidence are comparatively diffuse. This example complements the main paper by showing that LEMON can attribute evidence to non-image modalities when they provide direct clinical cues.

\end{document}